\documentclass[10pt,twocolumn,letterpaper]{article}

\usepackage[pagenumbers]{cvpr} % To force page numbers, e.g. for an arXiv version

\usepackage{graphicx}
\usepackage{amsmath}
\usepackage{amssymb}
\usepackage{booktabs}
\usepackage{amsmath,amsthm,amssymb,amsfonts}
\usepackage[ruled]{algorithm2e} %带竖线
\usepackage{enumitem} 
\usepackage{caption}
\usepackage{multirow}
\usepackage[pagebackref,breaklinks,colorlinks]{hyperref}

\newcommand{\xv}[0]{\ensuremath{\boldsymbol{x}} }

\usepackage[capitalize]{cleveref}
\crefname{section}{Sec.}{Secs.}
\Crefname{section}{Section}{Sections}
\Crefname{table}{Table}{Tables}
\crefname{table}{Tab.}{Tabs.}

\def\cvprPaperID{5297} % *** Enter the CVPR Paper ID here
\def\confName{CVPR}
\def\confYear{2023}

\begin{document}

%\title{ConZIC: Controllable Zero-shot Image Captioning in a Polishing Way}
\title{ConZIC: Controllable Zero-shot Image Captioning by Sampling-Based Polishing}

%\title{Zero-shot Controllable Image Captioning by sampling-based algorithm}
%\title{Teach CLIP to speek: \\
%A Zero-shot Controllable Image Captioning model}
% \title{Zero-shot diverse and controllable image captioning by looking around}
% \title{Tell what you want: controllable image captioning by looking around}

\author{Zequn Zeng\thanks{Equal contribution. \hspace{4mm}  \textdagger Corresponding authors}, Hao Zhang\footnotemark[1], Ruiying Lu, Dongsheng Wang, Bo Chen\footnotemark[2]\\
% \thanks{Corresponding authors}\\
National Key Laboratory of Radar Signal Processing, Xidian University, Xi’an, 710071, China\\
{\tt\small\{zzequn99, zhanghao\_xidian\}@163.com, bchen@mail.xidian.edu.cn}
% \authornote{*Corresponding authors}
\and
 Zhengjue Wang\\
State Key Laboratory of Integrated Service Networks, Xidian University, Xi’an, 710071, China\\
{\tt\small zhengjuewang@163.com}
}

\maketitle
%%%%%%%%% ABSTRACT
\begin{abstract}
Zero-shot capability has been considered as a new revolution of deep learning, letting machines work on tasks without curated training data.
%which let machines pre-trained on some dataset or task can imitate human to work well on the new one.
As a good start and the only existing outcome of zero-shot image captioning (IC), ZeroCap abandons supervised training and sequentially searches every word in the caption using the knowledge of large-scale pre-trained models.
Though effective, its autoregressive generation and gradient-directed searching mechanism limit the diversity of captions and inference speed, respectively.
Moreover, ZeroCap does not consider the controllability issue of zero-shot IC.
To move forward, we propose a framework for {\textbf{Con}}trollable {\textbf{Z}}ero-shot {\textbf{IC}}, named {\textbf{ConZIC}}.
The core of ConZIC is a novel sampling-based non-autoregressive language model named
Gibbs-BERT, which can generate and continuously polish every word.
Extensive quantitative and qualitative results demonstrate the superior performance of our proposed ConZIC for both zero-shot IC and controllable zero-shot IC. 
Especially, ConZIC achieves about 5$\times$ faster generation speed than ZeroCap, and about 1.5$\times$ higher diversity scores, with accurate generation given different control signals. Our code is available at 
\href{https://github.com/joeyz0z/ConZIC}{https://github.com/joeyz0z/ConZIC}.
\end{abstract}
\vspace{-6mm}

%%%%%%%%% BODY TEXT
\section{Introduction}
\label{sec:intro}

Image captioning (IC) is a visual-language task, which targets at automatically describing an image by generating a coherent sentence.
By performing supervised learning on human-annotated datasets, such as MS-COCO~\cite{lin2014microsoft}, many  methods~\cite{fang2022injecting,nguyen2022grit,mokady2021clipcap,hu2022scaling} have achieved impressive evaluation scores on metrics like BLEU~\cite{papineni2002bleu}, METEOR~\cite{banerjee2005meteor}, CIDERr~\cite{vedantam2015cider}, and SPICE~\cite{anderson2016spice}.
However, these methods still lag behind human capability of zero-shot IC.

\begin{figure}[!t]
  \centering
  \subfloat[Examples of zero-shot image captioning.]{\includegraphics[width=0.9\linewidth]{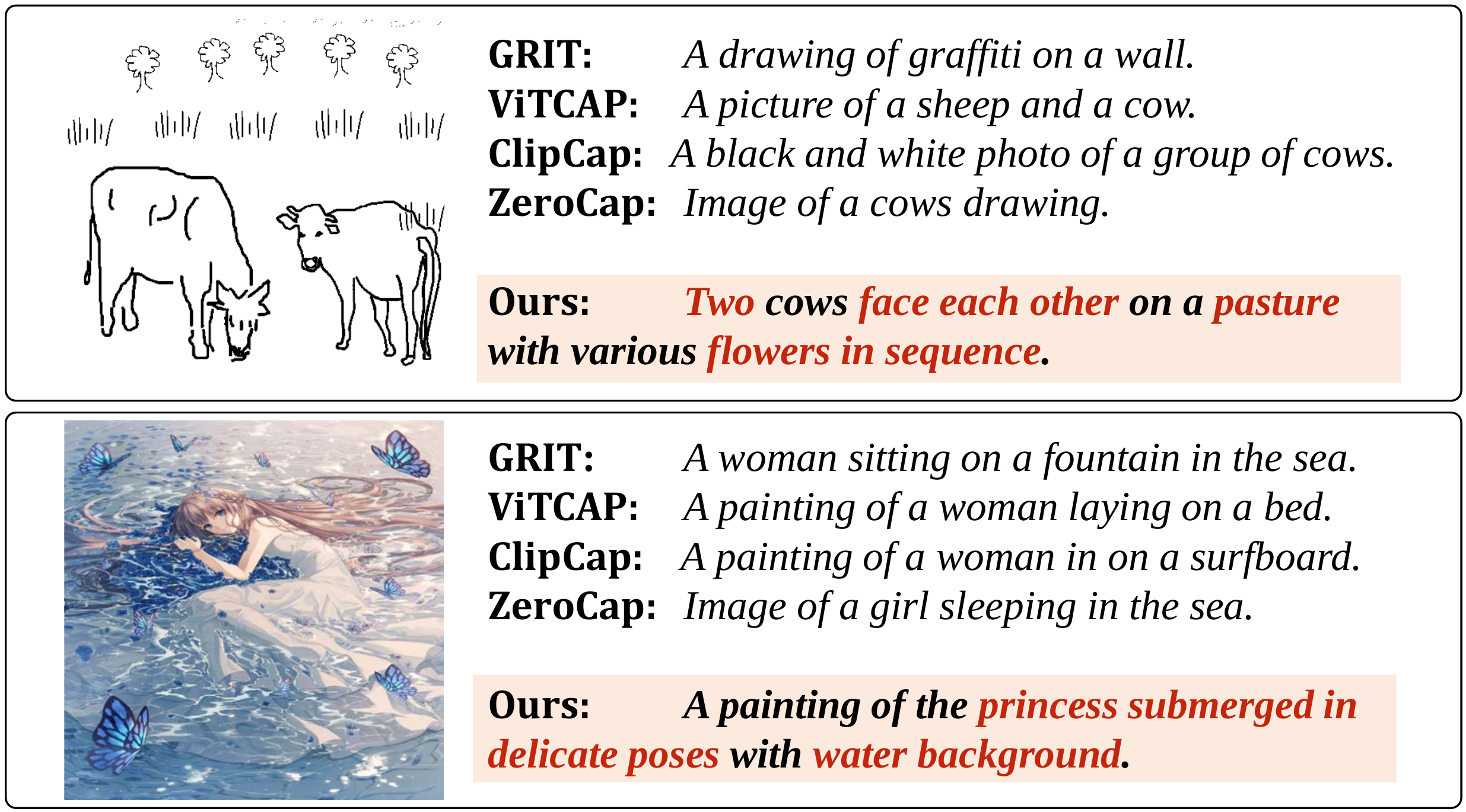}\label{fig1a}} 
  \quad
  \subfloat[Diversity of ConZIC.]{\includegraphics[width=0.9\linewidth]{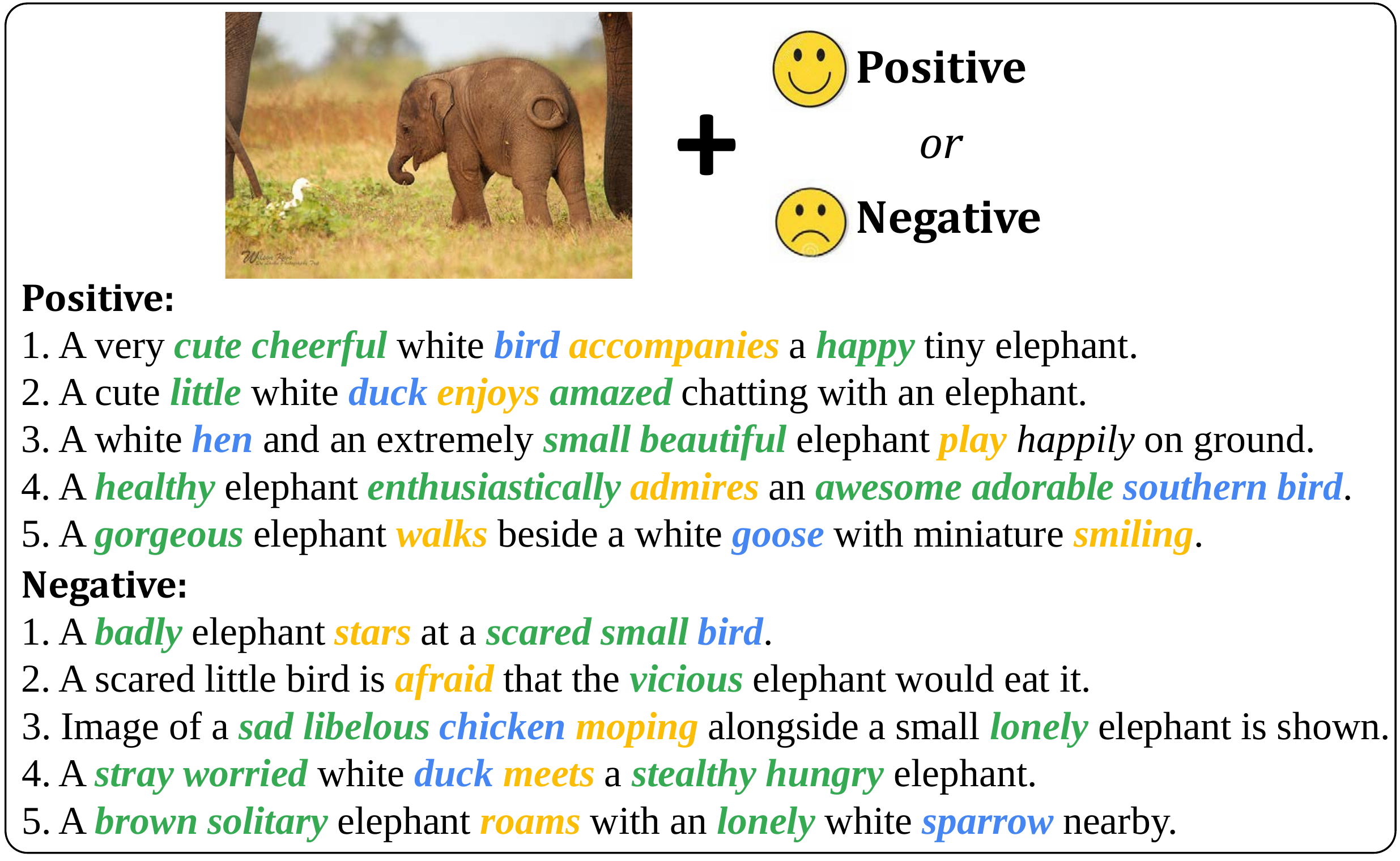}\label{fig1b}} 
  \vspace{-3mm}
  \caption{The highlights of our proposed method. (a) shows two examples of zero-shot image captioning on several SOTA methods. Specifically, GRIT ~\cite{nguyen2022grit} and ViTCAP ~\cite{fang2022injecting} are two supervised methods without pre-trained models. ClipCap ~\cite{mokady2021clipcap} is a supervised method using pre-trained CLIP. GRIT, ViTCAP, and CLIPCap are firstly trained on MSCOCO and then do testing. ZeroCap ~\cite{tewel2022zerocap} is the zero-shot method without any training. (b) shows the diversity of our proposed ConZIC, which manifests two aspects: {\textit{semantic}} (diverse words: different colors denoting different parts-of-speech) and {\textit{syntactic}} (diverse sentence patterns).}
  \vspace{-6mm}
  \label{fig1}
\end{figure}

Specifically, those supervised methods extremely rely on well-designed image-captions pairs.
However, it is likely impossible to construct a large enough dataset, including paired images and high-quality captions covering various styles/contents.
As a result, it is challenging for the machine to caption images that are outliers with respect to the training distribution, which is common in real applications (see examples in Fig.~\ref{fig1a}).
On the contrary, humans can perform IC without any specific training, \ie, realizing zero-shot IC.
Because humans can integrate what they see, \ie, the image, and what they know, \ie, the knowledge.

Recently, large-scale pretraining models have shown a strong capability of learning knowledge from super-large-scale data, showing great potential in various downstream tasks~\cite{radford2021learning, brown2020language, song2022clip,goh2021multimodal,petroni2019language}.
Equipped with the visual-language knowledge learned by CLIP~\cite{radford2021learning} and linguistic knowledge from GPT-2~\cite{radford2019language}, ZeroCap~\cite{tewel2022zerocap} is the first and the only zero-shot IC method, which proposes a searching-based strategy and is free of training on extra supervised data.
Specifically, ZeroCap searches the caption words one by one and from left to right, guided by CLIP-induced score for
image-text matching and GPT-2 word distribution for caption fluency.
ZeroCap is a good start and inspires us to explore how to search for the optimal caption in a better way.

\textit{i) More flexible.}
ZeroCap utilizes GPT-2 to perform left-to-right autoregressive generation. %This means the searching for the optimal word for current position only relies on previous generated words, not considering the full context.
Once a word is fixed, there is no chance to modify it when we move to the next position.
In other words, such generation order is not flexible enough to consider the full context information.

\textit{ii) More efficient.}
The searching at every position is realized by iteratively updating the parameters of GPT-2, which is time-consuming, as shown in Fig.~\ref{length_speed}.

\textit{iii) More diverse.}
IC is an open problem.
Given an image, different persons may have different visual attentions~\cite{chen2020say} and language describing styles~\cite{mathews2016senticap,gan2017stylenet,wang2019describing}, thus resulting in diverse descriptions. 
ZeroCap employs beam search to generate several candidate sentences, which, however, have similar syntactic patterns (see Appendix).

\textit{iv) More controllable.}
To endow captioning models with human-like controllability, \eg, sentiment, personality,
a recent surge of efforts ~\cite{chen2021human,mathews2016senticap,gan2017stylenet,deshpande2019fast} resort to introducing extra control signals as constraints of the generated captions, called Controllable IC.
However, controllable zero-shot IC has not been explored yet.

\begin{figure*}[!t]
	\centering 
	\includegraphics[width=151mm]{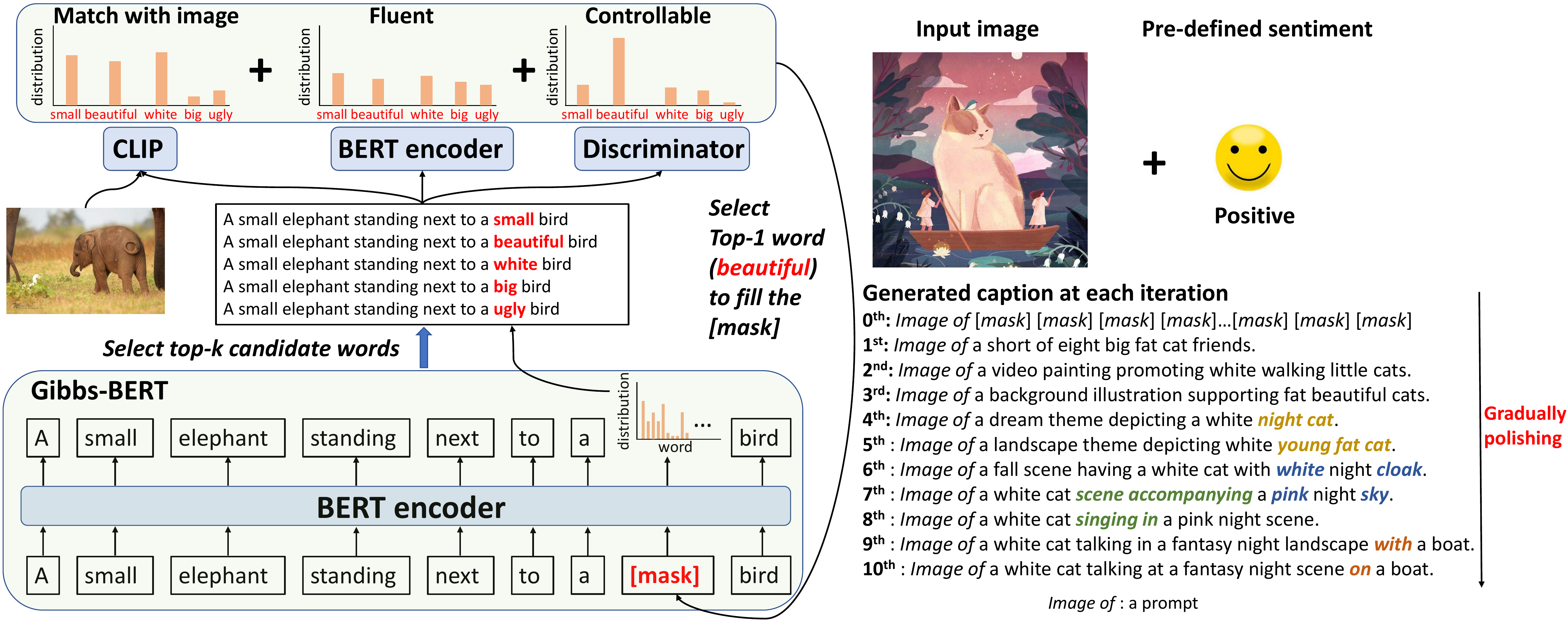}
	\caption{
	An overview of our approach. ConZIC starts from a prompt ``Image of'' and a [{\textit{mask}}] sequence and then iteratively updates the caption by sampling each word (see right). As an example (see left), ConZIC selects the word ``beautiful'' by considering image-matching score (in Sec.~\ref{image-text matching}), fluent score (in Sec.~\ref{Gibbs-BERT}), and controllable score (in Sec.~\ref{discriminator}) whose specific algorithm is in Algorithm \ref{alg::our method}. ConZIC can correct itself for zero-shot generation, as illustrated with the same color between two iterations at right. 
% 	The illustration of our proposed ConZIC for sampling one word at each iteration. Specifically, we firstly set it as [{\text{mask}}] token, and Gibbs-BERT can predict a word distribution over the vocabulary. Then we select top-$K$ candidate words forming $K$ sentences with other words. These sentences are fed into the discriminator for  controllin, and fed into CLIP with given images, to obtain three distributions over these $K$ candidate words. After integrating these three distributions, we select one word with the largest probability. Note that by deleting the discriminator, our proposed method can do zero-shot IC without controller. ConZIC starts from a noisy input like all [{\text{mask}}] tokens, and iteratively samples the captions in a polishing way. The color of words between two iterations illustrates that ConZIC can correct itself for zero-shot generation.
	}\label{Fig1:model}
\vspace{-5mm}
\end{figure*}

Bearing all these four-aspect concerns in mind, we propose a novel framework for controllable zero-shot IC, named ConZIC, as shown in Fig.~\ref{Fig1:model}.
Specifically, after analyzing the relationship between Gibbs sampling and masked language models (MLMs, currently we use BERT)~\cite{casella1992explaining,devlin2018bert,wang2019bert}, we firstly develop a new language model (LM) called Gibbs-BERT to realize the zero-shot IC by sampling-based search. Compared with autoregressive models, Gibbs-BERT has more a flexible generation order, bringing the self-correct capability by bidirectional attention with faster and more diverse generations. After integrating Gibbs-BERT with the CLIP that is used to evaluate the similarity between image and text, our proposed framework can perform zero-shot IC. By further introducing a task-specific discriminator for control signal into our framework, our proposed framework can perform controllable zero-shot IC.

The main contributions of this paper are:
\vspace{-2mm}
\begin{itemize}[leftmargin=10pt]
	\setlength{\itemsep}{0.1pt}
	\setlength{\parsep}{0pt}
	\setlength{\parskip}{1pt}

    \item We propose to solve the controllable zero-shot IC task in a polishing way. By combining Gibbs sampling with a MLM, we can randomly initialize the caption and then polish every word based on the full context (bidirectional information) in the caption.
    
    \item ConZIC is free of parameter updates, achieving about 5$\times$ faster generation speed than the SOTA method, ZeroCap. 

    \item Equipped with Gibbs-BERT, ConZIC can perform flexible searching, thus generating sentences with higher diversity, as shown in Table. ~\ref{Table:MSCOCO}. 

    \item To the best of our knowledge, ConZIC is the first controllable zero-shot IC method. Four classes of controllable signals, including length, infilling, styles, and parts-of-speech, are evaluated in our experiments.

\end{itemize}

\vspace{-5mm}
\section{Related work}
\label{sec:related_work}
\subsection{Supervised Image captioning}

To generate text description given an input image,
traditional image captioning (IC) often relies on curated image-caption pairs to train an encoder-decoder model.
For example, some early attempts~\cite{vinyals2015show, donahue2015long,xu2015show,gu2017empirical} construct 
CNN-based encoder to extract visual features and RNN/LSTM-based decoder to generate output sentence.
For better visual understanding, some methods~\cite{anderson2018bottom, cornia2020meshed, huang2019attention, huang2019adaptively, qin2019look, wang2020show, kuo2022beyond} employ an object detector to extract attentive image regions.
To encourage more interactions between two modalities, attention mechanism~\cite{cornia2020meshed,huang2019attention, nguyen2022grit, pan2020x, schwartz2019simple, schwartz2017high} and graph neural
network~\cite{yang2019auto,yao2018exploring} have been widely adopted.

Recently, a series of large-scale visual-language pre-training models ~\cite{li2020oscar,zhang2021multi,zhou2020unified,hu2022scaling,jia2021scaling, radford2021learning} have been built, showing remarkable performance in various downstream tasks, IC included. 
Equipped with these pre-trained models, \cite{hu2022scaling,zhang2021vinvl, su2022language,mokady2021clipcap,fang2022injecting} have shown SOTA performances.
However, these methods still need supervised fine-tuning on human-annotated datasets, such as MS-COCO.

\vspace{-1mm}

\subsection{Zero-shot Image Captioning}
\vspace{-1mm}

Zero-shot capability with large-scale pretrained models has drawn much attention in a wide range of research fields in computer vision and natural language processing~\cite{radford2021learning,zhai2022lit,bansal2018zero,pourpanah2022review,radford2019language,brown2020language}, showing great potential of transferring knowledge to tasks without supervised training data.
However, for IC, the zero-shot ability is under-explored. 
To the best of our knowledge, the only achievement was made by Tewel \etal. in~\cite{tewel2022zerocap}, where a method called ZeroCap is proposed. 
Additionally, several approaches address the task of describing novel objects which are not present in paired image-sentence training data by incorporating external unsupervised data ~\cite{agrawal2019nocaps, hendricks2016deep, venugopalan2017captioning} or object tagger ~\cite{anderson2016guided, feng2020cascaded, hu2021vivo, li2019pointing, lu2018neural}, namely novel object image captioning (NOIC). 
In some cases, NOIC is also called zero-shot IC, but is very different from what we study in this work. A detailed comparison is in Appendix ~\ref{appendixA}.

Specifically, ZeroCap uses a pre-trained GPT-2.
% Then, for the generation at each position, they update the context cache by maximizing the image-text matching score measured by CLIP.
Then, for the generation at each position, they update the context cache by minimizing the image-text matching loss measured by CLIP at inference stage.
Note that CLIP is a visual-language pretraining model trained on an automatically collected noisy web-scale dataset, rather than human-annotated IC datasets.
As a result, ZeroCap realizes zero-shot IC by gradient-directed searching without training.

However, due to the autoregressive nature, 
searching for the word at current position only considers the information from the left side, not the full context.
Besides, the autoregressive nature tends to bring mode collapse problem~\cite{xiao2022survey}, resulting in captions with less diversity.
Moreover, the time-cost of iterative gradient-update is high, especially for long captions.
Further, ZeroCap has not been considered for the task of controllable zero-shot IC.

\subsection{Diversity and Controllability}

Diversity~\cite{wang2019describing, vijayakumar2018diverse, wang2017diverse} and controllability~\cite{chen2020say, deshpande2019fast, gan2017semantic,guo2019mscap ,gan2017stylenet, cornia2019show,chunseong2017attend,deng2020length,kim2019dense} are two important properties that have drawn much attention in previous IC researches.

Recent findings~\cite{chen2022learning} show that the captions generated by supervised methods tend to be biased toward the ``average” caption, capturing the most general linguistic patterns and words in the training corpus, \ie, the so-called mode collapse problem.
In other words, semantically, ``diversity" refers to various words, and syntactically, ``diversity" refers to abundant sentence patterns.
%Generally, ``diversity" often refers to two aspects: the diversity of sentence patterns and the diversity of generated words.
Without the limitation of supervised data and using the knowledge from CLIP, ZeroCap increases the vocabulary size of generated captions.
Nevertheless, its autoregressive nature brings a phenomenon that the candidate sentences for a given image often have similar syntactic patterns, \ie, less syntactical diversity.

To imitate human controllability, lots of works have been made to control the caption generation by introducing control signals into the supervised training process, such as the subjective signals like sentiments~\cite{gan2017semantic,guo2019mscap,zhao2020memcap}, emotions~\cite{gan2017stylenet,mathews2018semstyle}, personality~\cite{chunseong2017attend,shuster2019engaging} and the objective signals like length level~\cite{deng2020length}, parts-of-speech~\cite{deshpande2019fast}, object region~\cite{cornia2019show,lindh2020language}, and visual relation~\cite{kim2019dense}.
However, how to introduce control signals without training and realize controllable zero-shot IC has not been explored yet.

\vspace{-2mm}
\section{Method}
\label{Method}
\vspace{-1mm}
\subsection{Framework of ConZIC}

Given an image $I$, zero-shot image captioning (IC) aims to generate a linguistic description $\xv_{<1,n>}$ containing $n$ words, without training on the supervised database. 
This process can be formalized as searching $\xv_{<1,n>}$ by maximizing the data likelihood $p(\xv_{<1,n>}|I)$.

To further consider the influence of a control signal $C$, controllable zero-shot IC focuses on searching $\xv_{<1,n>}$ by maximizing $p(\xv_{<1,n>}|I,C)$. 
According to the Bayes rule, the log data likelihood can be derived as:
\begin{align}
\label{eq: decompose_1}
         {}&{} \log p(\xv_{<1,n>}|I,C)  \notag \\
\propto  {}&{} \log p(\xv_{<1,n>},I,C) \\
= {}&{} \log p(I|\xv_{<1,n>}) + \log p(C|\xv_{<1,n>})+ \log p(\xv_{<1,n>}) \notag,
\vspace{-2mm}
\end{align}
which implies three basic rules to guide the searching process, realized by three modules, respectively.
Specifically, \textit{i)} a language model (LM), evaluating $p(\xv_{<1,n>})$, helps with searching for captions with high-level fluency;
\textit{ii)} a matching network, measuring the similarity between the input image and the generated caption, \ie, $p(I|\xv_{<1,n>})$, helps with searching for captions highly related to the input image;
and \textit{ii)} a discriminator, measuring $p(C|\xv_{<1,n>})$, helps with searching for captions that meet the control signal.  

These three modules constitute our proposed controllable zero-shot IC framework, ConZIC, which will be further introduced in the following subsections.
ConZIC tries to solve the controllable zero-shot IC problem by iteratively polishing the words at every position.

\subsection{Sampling-based language model for $p(\xv_{<1,n>})$}
\label{Gibbs-BERT}

To model $p(\xv_{<1,n>})$, existing IC methods (including zero-shot and supervised ones) often adopt sequential autoregressive generations, as:
\begin{equation}
p(\xv_{<1,n>}) 
= p(x_n|\xv_{<n}) \cdots p(x_2|x_1)p(x_1).
\end{equation}
However, such autoregressive generation often results in issues such as sequential error accumulation and lack of diversity~\cite{chen2022learning,xiao2022survey}.
Further, for zero-shot IC, the sequential searching-order is lack of flexible.
See related work for more detailed discussions. 
To move beyond, inspired by our analysis of the relation between Gibbs sampling and the design of masked language models (MLMs), we develop a sampling-based LM for $p(\xv_{<1,n>})$.

Specifically, Gibbs sampling is a Markov chain Monte Carlo (MCMC) algorithm, which aims to collect samples from the joint data distribution $p(\xv_{<1,n>})$ by sampling each variable $x_i$ (word in our case) iteratively from its conditional 
probability $p(x_i|\xv_{-i})$, where $\xv_{-i}$ denotes all other random variables in $p(\xv_{<1,n>})$ except $x_i$.
In practice, Gibbs sampling brings flexible sampling orders like
\vspace{-2mm}
\begin{align}
\label{eq: decompose_2}
\small
p(x_n|\xv_{-n})& \rightarrow p(x_{n-1}|\xv_{-(n-1)}) \rightarrow \cdots \rightarrow p(x_1|\xv_{-1}) \notag \\
p(x_1|\xv_{-1})& \rightarrow p(x_2|\xv_{-2}) \rightarrow\cdots \rightarrow p(x_n|\xv_{-n}) \\
p(x_t|\xv_{-t})& \rightarrow \cdots \rightarrow p(x_j|\xv_{-j}). \notag
\end{align}
Such flexible order gives Gibbs sampling the ability to walk out the collapsed modes (a key problem of lacking diversity in IC~\cite{chen2022learning}), resulting in more diverse generations.  

From another view to analyze Eq.~\ref{eq: decompose_2}, each item is associated with the learning of MLMs.
Specifically, given a sentence, MLMs set several words as the [{\textit{MASK}}] denoted by $\xv_{\mathbb{M}}$, and then use other words $\xv_{-\mathbb{M}}$ to predict these masked words.
Mathematically, the target of MLMs is to learn the conditional distribution $p(\xv_{\mathbb{M}}|\xv_{-\mathbb{M}})$ from the corpus.
Therefore, if we just set $i$-th word $x_i$ as the [{\textit{MASK}}], MLMs and Gibbs sampling are equivalent to predict $p(x_i|\xv_{-i})$.
Currently, we use BERT as MLMs and therefore we call this new LM as Gibbs-BERT to model $p(\xv_{<1,n>})$. 

The specific algorithm of Gibbs-BERT for sampling a sentence $\xv_{<1,n>}$ from $p(\xv_{<1,n>})$ is shown in Algorithm 2 in Appendix ~\ref{appendixB}.
After randomly choosing the generation order, Gibbs-BERT starts from a full noisy sentence ($\eg$, all [{\textit{MASK}}] tokens).
At each iteration, Gibbs-BERT progressively samples each word by putting [{\textit{MASK}}] at this position and then selecting the top-$1$ word from the predicted word distribution over the vocabulary by BERT.
The result of $t$-th iteration is the initialization of the $(t+1)$-th iteration.

% \vspace{-2mm}
\subsection{Image-text matching network for $p(I|\xv_{<1,n>})$}
\label{image-text matching}
To make the generated caption highly related to the image, our framework needs a matching network that can measure the similarity between images and texts.
Recently, pre-trained on the sufficiently large-scale image-text pairs, CLIP~\cite{radford2021learning} learns abundant world knowledge for measuring their similarity. 
Therefore, we introduce the pre-trained CLIP into our framework for modeling $p(I|\xv_{<1,n>})$.

Specifically, when sampling each word as $p(x_i|\xv_{-i};I)$, Gibbs-BERT firstly provides top-$K$ candidate words according to its predicted word distribution over the vocabulary.
Then we replace the [{\textit{MASK}}] token for $i$-th position with these $K$ candidate words, forming $K$ candidate sentences $\{s_{k}=(x_1,...,x_{ik},...,x_n)\}_{k=1}^{K}; x_{ik}=$[\textit{MASK}].
The CLIP matching score $p(I|s_k)$ can be computed as $CLIP(s_k,I)$, where a higher score represents that image and text are better aligned.
Using the Softmax, we obtain a predicted distribution over these $K$ candidate words as
\begin{equation}
\label{eq: clip score}
\small
p(I|\{s_k\}_{k=1}^K) \propto Softmax[CLIP(s_k,I)].
\end{equation}
According to Eq.~\ref{eq: clip score}, we select the top-$1$ word (largest probability) as $x_i$, forming the sentence with other words $\xv_{-i}$.

Up to now, our framework has already realized the zero-shot IC without the control signal.
Next, we will introduce how to integrate a discriminator $p(C|\xv_{<1,n>})$ of the control signal $C$ for controllable zero-shot IC.

\begin{algorithm}[!t]
\caption{Algorithm of our proposed ConZIC.} % 名称
\label{alg::our method}
\KwData{initial caption:\small{$\xv_{<1,n>}^0=(x_1^0,...,x_n^0)$}\;
iterations=\small{$T$}, candidates=\small{$K$}\; position sequence \small{$P=\textit{Shuffle}([1,...,n])$\;}}
\KwResult{the final caption: \small{$\xv_{<1,n>}^T=(x_1^T,...,x_n^T)$;}}
\For{iteration \small{$t \in [1,...,T]$}}
{ 
    state: \small{$\xv_{<1,n>}^{t-1}=(x_1^{t-1},...,x_n^{t-1})$}\; 
    \For{position $i \in P$}
    {
        1. Replace \small{$x_i^{t-1}$} with \small{[\textit{MASK}]};\\
        2. Predict the word distribution over vocabulary by Gibbs-BERT: 
        \small{$p(x_i|\xv_{-i}^{t-1})$}\;
        3. Select top-$K$ candidate words 
        \small{$\{x_{ik}^{t}\}_{k=1}$} by \small{$p(x_i|\xv_{-i}^{t-1})$}, whose probability is \small{$p^{Bert}_k$}\;
        4. Get $K$ candidate sentences \small{$\{s_k\}_{k=1}^K$}: \\
        \small{$(x_1^{t-1},...,,x_{i-1}^{t-1},x_{ik}^{t},x_{i+1}^{t-1},...,x_n^{t-1})_{k=1}^K$};\\
        %$p_{flu}\propto p(\{x_{ik}^{t}\}_{k=1}^K)$; \\
        5. Compute the CLIP and classifier score for \small{$\{s_k\}_{k=1}^K$} by Eq.~\ref{eq: clip score} and \ref{eq: discriminator score}: \small{$p^{Clip}_k$ and $p^{Cls}_k$}. \\       
        6. Select \small{$x_i^{t}$} with largest probability by \small{$\alpha p^{Bert}_k + \beta p^{Clip}_k + \gamma p^{Cls}_k$}\;
        7. Replace \small{$x_i^{t-1}$} with \small{$x_i^{t}$}\;
    }
    state: \small{$\xv_{<1,n>}^{t}=(x_1^{t},...,x_n^{t})$}\; 
}

\end{algorithm}

\subsection{Discriminator for control signal $p(C|\xv_{<1,n>})$}
\label{discriminator}
As for controllable IC, we need to generate text related to both image and given control signal $C$.
For some types of control signals, like sentiment or parts-of-speech (POS), we need an extra discriminator $p(C|\xv_{<1,n>})$ to evaluate the correlation between caption and control signals.
Specifically, similar to what we do for $p(I|\xv_{<1,n>})$, after selecting top-$K$ sentences $s_k$ by Gibbs-BERT, we use a pre-trained classifier with Softmax function to model $p(C|\{ s_k\}_{k=1}^K)$ as
\begin{equation}
\small
\label{eq: discriminator score}
p(C|\{ s_k\}_{k=1}^K) \propto Softmax[Classifier(s_k)].
\end{equation}
The classifier is different for different tasks, which will be detailed in the experiments.

\subsection{Overall algorithm}
\label{overall framework}
The overall algorithm of the framework for controllable zero-shot IC is shown in Algorithm \ref{alg::our method}. 
We firstly need to initial the caption (currently we use all [$MASK$] tokens) and set several hyper-parameters.
Starting from the output of the previous iteration, for each position $i$ of this iteration, Gibbs-BERT firstly provides top-$K$ candidate words, forming $K$ sentences with other words denoted as $\{s_{k}\}_{k=1}^K$.
Then, according to Eq.~\ref{eq: clip score} and Eq.~\ref{eq: discriminator score}, we can obtain the text-image and text-control matching scores $p(I|\xv_{<1,n>})$ and $p(C|\xv_{<1,n>})$.
After integrating these two distributions with Gibbs-BERT predicted distributions $p(x_i|\xv_{-i})$ by trade-off parameters $\{\alpha, \beta, \gamma\}$, we can get a final distribution, from which we select the word with the largest probability as $x_i$.

There are three points about our proposed framework that we need to clarify.
\textit{i)} deleting the item $p(C|\xv_{<1,n>})$, our framework can do standard zero-shot IC (without control signal).
\textit{ii)} for some tasks of controllable IC, such as length control, there is no need to use $p(C|\xv_{<1,n>})$, whose details are in experiments.
\textit{iii)} our framework is free of the specific modules.
As our future exploration, we will study whether better pre-trained models, like using RoBERTa ~\cite{liu2019roberta} to replace BERT and ALIGN ~\cite{jia2021scaling} to replace CLIP, can further improve the performance of ConZIC.

\begin{table*}[htbp!]
\vspace{-3mm}
\centering
\resizebox{0.9\textwidth}{!}{
    \begin{tabular}{c|ccccc|c|cccc}
        \toprule[2pt]
        &\multicolumn{6}{c|}{Accuracy} & \multicolumn{4}{c}{Diversity} \\\hline
        \multirow{2}{*}{Metrics}&\multicolumn{5}{c|}{Supervised} & \multicolumn{1}{c|}{Unsupervised} &  \\
        &B-4($\uparrow$)&M($\uparrow$)& C($\uparrow$) & S($\uparrow$) & RefCLIP-S($\uparrow$) & CLIP-S($\uparrow$) &Vocab ($\uparrow$) &S-C($\uparrow$) &Div-1($\uparrow$) &Div-2($\uparrow$) \\\hline
        \multicolumn{11}{c}{Supervised Methods} \\\hline
        ClipCap~\cite{mokady2021clipcap} &32.15 &27.1 &108.35 &20.12 &0.81 & 0.77 &1650 &- &- &- \\
        MAGIC~\cite{su2022language} &12.90 &17.22 &48.33 &10.92 &0.77 & 0.74 &1765 &- &- &-\\
        CLIP-VL~\cite{shen2021much} &40.2 &29.7 &134.2 &23.8 &0.82 & 0.77 &2464 &- &- &- \\
        ViTCAP~\cite{fang2022injecting} &41.2 &30.1 &138.1 &24.1 &0.80 &0.73  &1173 &- &- &- \\
        GRIT~\cite{nguyen2022grit} &42.4 &30.6 &144.2 &24.3 &0.82 & 0.77 &1049 &- &- &- \\
        VinVL~\cite{zhang2021vinvl} &41.0 &31.1 &140.9 &25.2 &\textbf{0.83} & 0.78 &1125 &- &- &- \\
        % Oscar~\cite{} &41.7 &30.6 &140.0 &24.5 &-  &- &- &- &- &- \\
        LEMON~\cite{hu2022scaling} &\textbf{42.6} &\textbf{31.4}
        &\textbf{145.5} &\textbf{25.5} &-  &- &- &- &- &- \\\hline
        \multicolumn{11}{c}{Supervised and Diversity-based Methods} \\\hline
        Div-BS~\cite{wang2017diverse} &32.5 &25.5 &103.4 &18.7 &-  &- &- &- &0.20 &0.25\\
        AG-CVAE~\cite{vijayakumar2018diverse} &31.1 &24.5 &100.1 &17.9 &- & - &- &- &0.23 &0.32\\
        POS~\cite{deshpande2019fast} &31.6 &25.5 &104.5 &18.8 &-  & - &- &- &0.24 &0.35\\
        ASG2Caption~\cite{chen2020say} &31.6 &25.5 &104.5 &18.8 &-  & - &- &0.76 &0.43 &0.56\\\hline
        \multicolumn{11}{c}{Zero Shot Methods} \\\hline
        ZeroCap~\cite{tewel2022zerocap} &2.60 &11.50 &14.60 &5.50 &0.79 &0.87 &8681 &0.63 &0.31 &0.45 \\
        \textbf{Ours (sequential)} &1.31 &11.54 &12.84 &5.17 &\textbf{0.83} &\textbf{1.01} &9566 &0.63 &0.40 &0.56  \\
        \textbf{Ours (shuffle)} &1.29 &11.23 &13.26 &5.01 &\textbf{0.83} &0.99 &\textbf{15462} &\textbf{0.95} &\textbf{0.62} &\textbf{0.87}  \\
        \bottomrule[2pt]
    \end{tabular}}
% \end{center}
\vspace{-3mm}
\caption{\small{Performance compared with SOTA methods on MSCOCO dataset. The baselines can be divided into three parts, supervised, supervised diversity-based methods and zero-shot methods.
Ours (sequential) is our framework with sequential generated order, while ours (shuffle) is randomly shuffled generated order. For accuracy metrics, we report the CLIP re-ranked best-1 performance among all iterations. To compute the diversity metrics which need multiple captions, Ours (sequential) is computed on 5 captions in last 5 iteration steps while Ours (shuffle) first randomly sample 5 generation orders, and then select the last-one caption.}}
\label{Table:MSCOCO}
\vspace{-2mm}
\end{table*}

% \begin{figure*}[!t]
% 	\centering 
% 	\includegraphics[width=160mm]{length_results.pdf}
% 	\vspace{-3mm}
% 	\caption{
% % 	Performance-Iterations curse of our model with four length size, $\ie$ 5, 8, 12, 15 and zero-shot image captioning baseline Zerocap, on MSCOCO dataset.
% % 	(a) X-axis is the iterations, Y-axis is the image relatedness metric CLIPScore.(b) X-axis is the iterations, Y-axis is the semantic similarity metric with human annotations, $\ie$ RefCLIPScore. Particularly, Zerocap is a non iterative method, and thus its curse is a dotted line parallel to X-axis.
%     Semantic relatedness metrics and inference time consuming on different sentence length compared with Zerocap. Zerocap is a one-pass method without iterative update shown in blue dotted line parallel to X-axis in (a) and (b). (a) X-axis is the iterations, Y-axis is the image relatedness metric CLIPScore. (b) X-axis is the iterations, Y-axis is the semantic similarity metric with human annotations, $\ie$ RefCLIPScore. (c) X-axis is the sentence length with 15 iterations, Y-axis is average inference time per image. ConZIC achieves better accuracy performance and lower time consuming compared with Zerocap.
% 	}
% 	\label{Fig2:length}
% 	\vspace{-4mm}
% \end{figure*}

\begin{figure*}[!t]
  \centering
  \subfloat[Comparision on CLIPScore.]{\includegraphics[width=0.3\linewidth]{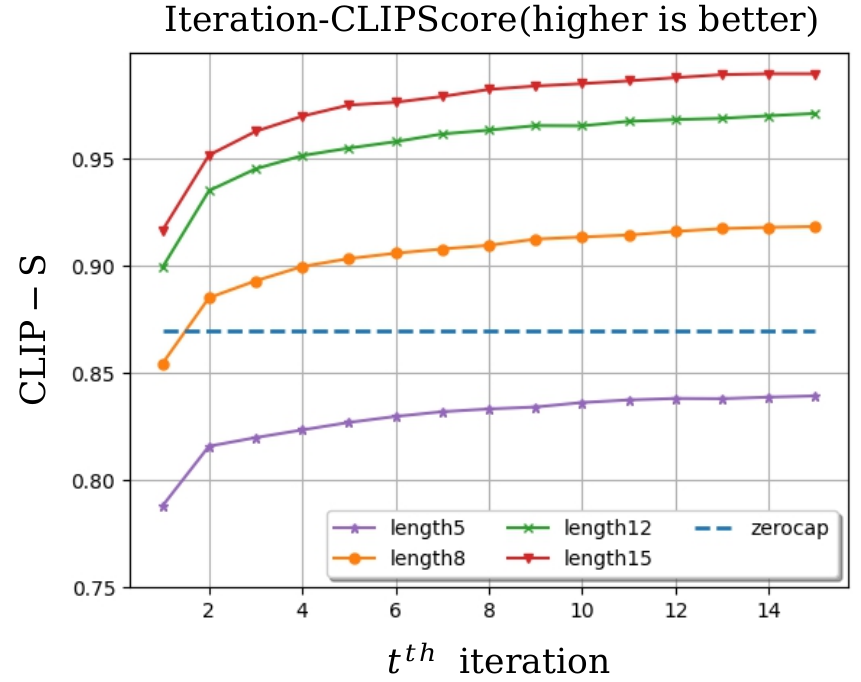}\label{length_clipscore}} 
  \subfloat[Comparision on RefCLIPScore.]{\includegraphics[width=0.3\linewidth]{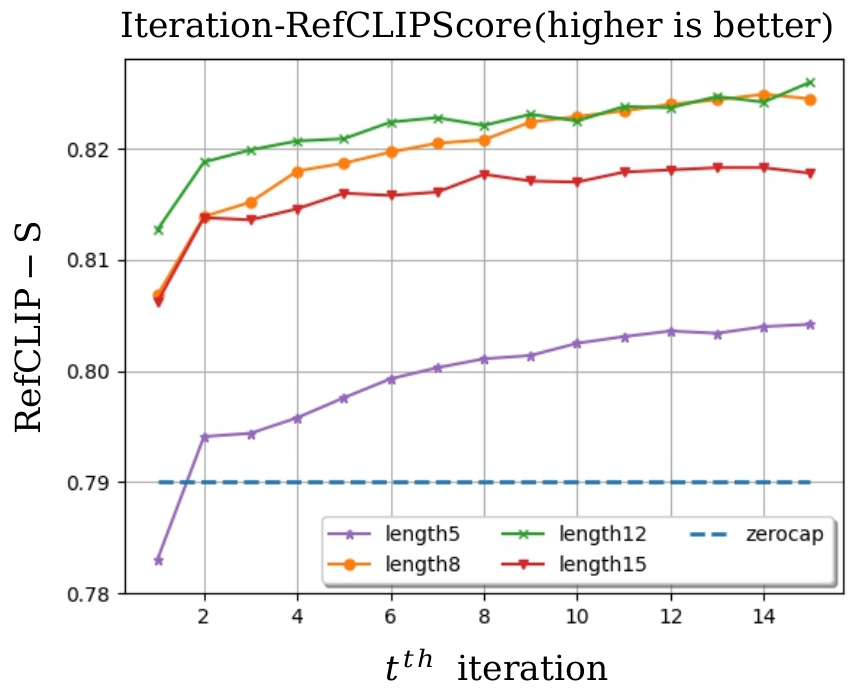}\label{length_refclipscore}} 
  \quad
  \subfloat[Comparision on Time-consuming.]{\includegraphics[width=0.32\linewidth]{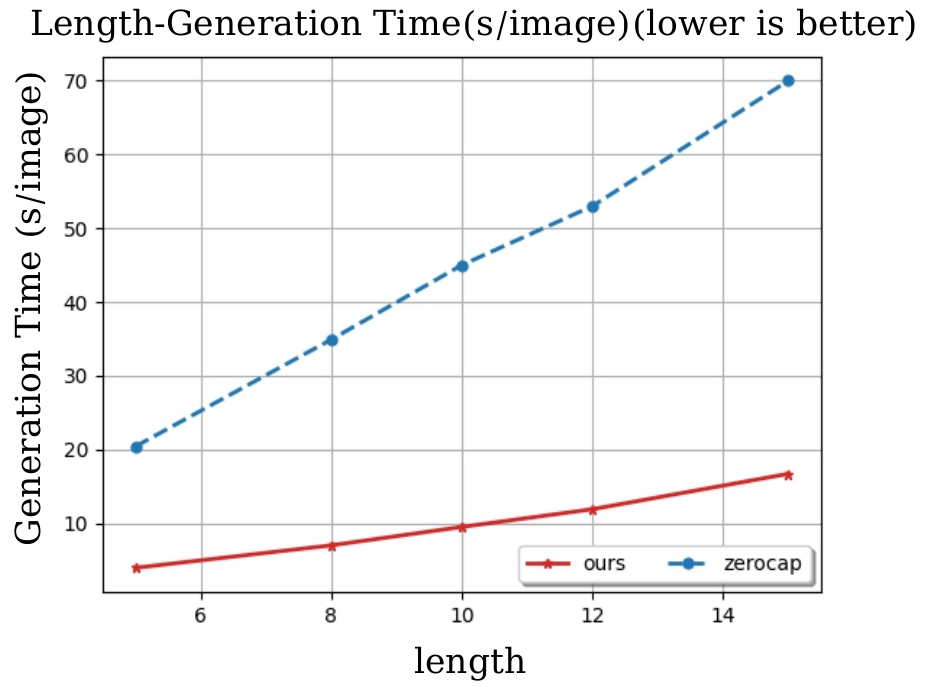}\label{length_speed}} 
  \vspace{-3mm}
  \caption{(a) and (b): The change of CLIPScore and RefCLIPScore with the iteration steps under different lengths of captions. (c): The generation speed with different lengths. %ZeroCap needs iteration for generating each word. 
  %Here, we just follow the best setting in the paper without tuning it. 
  Our framework achieves better accuracy and faster generation speed compared with Zerocap.
  }
  \vspace{-4mm}
  \label{Fig2:length}
\end{figure*}

% \begin{table}[htbp!]
% \begin{center}
%   \resizebox{0.45\textwidth}{!}{
%   \begin{tabular}{c|cccc}
%     \toprule[1.5pt]
%     Methods & B-1($\uparrow$) & M($\uparrow$) & C($\uparrow$) &CLIP-S($\uparrow$)\\ \hline
%     MAGIC~\cite{su2022language} &21.88 &11.77 &13.00 &0.66  \\
%     GRIT ~\cite{nguyen2022grit} &17.84 &\textbf{26.62} &17.84 &0.68\\
%     Zerocap~\cite{tewel2022zerocap} &27.08 &20.67 &21.11 &0.86 \\\hline
%     \textbf{Ours} &\textbf{39.61} &20.71 &\textbf{34.43} &\textbf{0.88} \\
%     \bottomrule[1.5pt]
%   \end{tabular}}
%   \end{center}
%   \vspace{-6mm}
%   \caption{\small{Performance on SketchyCOCO caption dataset.}}
%     \label{Table:sketchyCOCO}
%     \vspace{-3mm}
% \end{table}

\begin{table}[htbp!]
\begin{center}
  \resizebox{0.45\textwidth}{!}{
  \begin{tabular}{c|c|cccc}
    \toprule[1.5pt]
     \multicolumn{2}{c|}{Methods} & B-1($\uparrow$) & M($\uparrow$) & C($\uparrow$) &CLIP-S($\uparrow$)\\ \hline
    \multirow{3}*{Supervised} &MAGIC~\cite{su2022language} &21.88 &11.77 &13.00 &0.66  \\
    &ViTCAP~\cite{fang2022injecting} &27.69 &17.58 &22.29 &0.63  \\
    &GRIT ~\cite{nguyen2022grit} &17.84 &\textbf{26.62} &17.84 &0.68\\ \hline
    \multirow{2}*{Zero Shot} &ZeroCap~\cite{tewel2022zerocap} &27.08 &20.67 &21.11 &0.86 \\
    &\textbf{Ours} &\textbf{39.61} &20.71 &\textbf{34.43} &\textbf{0.88} \\
    \bottomrule[1.5pt]
  \end{tabular}}
  \end{center}
  \vspace{-6mm}
  \caption{\small{Performance on SketchyCOCO caption dataset.}}
    \label{Table:sketchyCOCO}
    \vspace{-5mm}
\end{table}

\section{Experiments}
\subsection{Datasets}
\textbf{MSCOCO caption}~\cite{lin2014microsoft}:
MSCOCO caption is a large IC dataset. 
we use the ‘Karpathy’ splits ~\cite{karpathy2015deep} that have been used extensively for reporting results in most of prior works. 
This split contains 113,287 training and validation images, and 5,000 for testing.
Each image contains five captions.

\textbf{SentiCap}~\cite{mathews2016senticap}: 
SentiCap is a sentiment IC dataset developed from the MSCOCO dataset.
Each image is labeled by 3 positive and/or 3 negative sentiment captions. 
As a result, the positive (negative) set contains 998 (673) and 997 (503) images for training (testing), respectively. 

\textbf{FlickrStyle10k}~\cite{gan2017stylenet}: 
FlickrStyle10k contains 10,000 Flickr images with stylized captions, where only 7,000 images are public.
Each image contains 5, 1, and 1 captions for factual (no specific style), humorous, and romantic styles, respectively. 
Following ~\cite{guo2019mscap}, we randomly select 6,000 and 1,000 of them to construct the training and testing sets.

\textbf{SketchyCOCO caption}~\cite{gao2020sketchycoco}: 
SketchyCOCO is constructed by collecting instance freehand sketches covering 3 background and 14 foreground classes. 
However, SketchyCOCO is not for IC since it only has the classification label.
To help us evaluate the performance of IC on sketch-style images quantitatively, we construct a small benchmark based on SketchyCOCO, through the following steps: \textit{i)} randomly sample 100 sketch images for each foreground class. \textit{ii)} label them with a simple prompt, $\ie$ ``A drawing of a [CLASS]'', where [CLASS] is the class name. For example, a cat image is labeled as ``A drawing of a cat.''. More details can be seen in Appendix ~\ref{appendixC}.

% \begin{table*}[htbp!]
% \begin{center}
%   \resizebox{0.9\textwidth}{!}{
%   \begin{tabular}{c|cccc|cccc}
%     \toprule[2pt]
%     & \multicolumn{4}{c|}{Romantic} & \multicolumn{4}{c}{Humorous} \\
%     Methods & B-3($\uparrow$) &  M($\uparrow$) & CLIP-S($\uparrow$) &Acc($\uparrow$) &B-3($\uparrow$) &M($\uparrow$) & CLIP-S($\uparrow$)  &Acc($\uparrow$)\\ \hline
%     StyleNet ~\cite{gan2017stylenet} &1.5 &4.5 &- &37.8 &0.90  & 4.30 &- &41.9 \\
%     MSCap ~\cite{guo2019mscap} &2.0 &5.4 &- &88.7 &1.90 &5.30 &- &91.3  \\
%     MemCap ~\cite{zhao2020memcap} &\textbf{4.0} &\textbf{7.7} &- &91.7 &\textbf{4.0}  &\textbf{7.20} &- &\textbf{97.1} \\\hline
%     \textbf{Ours} &1.2 &6.1 &\textbf{1.02} &\textbf{96.3} &1.2 &6.1 &\textbf{1.02} &91.4  \\
%     \bottomrule[2pt]
%   \end{tabular}}
%   \end{center}
%   \vspace{-5mm}
%   \caption{Stylized image captioning($\ie$ romantic, humorous) performance comparisons on the Flickstyle8k dataset.}
% \label{Table:flickstyle}
% \end{table*}

\subsection{Implementation Details}
As described in Sec.~\ref{Method}, all the experiments are executed based on frozen pre-trained models without any fine-tuning.
Specifically, we choose CLIP-ViT-B/32 as our image-text matching network and BERT-Base as our LM.
As for different controllable IC tasks, we select corresponding discriminators whose details are introduced in Sec.~\ref{evaluation on CIC tasks}.

In Sec.~\ref{Evaluation on Accuracy and Diversity}, we first evaluate the performance on standard IC without control signals. 
Then in Sec.~\ref{evaluation on CIC tasks}, we explore the controllability of our method on 4 controllable IC tasks including length, infilling, style, and parts-of-speech (POS).
Finally, in Sec.~\ref{Evaluation on Time-consuming}, we study the speed of generation.
$K,T,\alpha,\beta$ are set as 200, 15, 0.02, and 2 among all experiments.
For MSCOCO captions and SketchyCOCO captions, we set sentence lengths $n$ as 12 and 5, respectively.
As for stylized IC and POS controlled IC, we set $\gamma$ as 5.
All experiments are conducted on a single RTX3090 GPU.

\subsection{Evaluation on Accuracy and Diversity}
\label{Evaluation on Accuracy and Diversity}
We first evaluate the accuracy and diversity of our framework based on standard IC task (without control).

{\bf{Evaluation Metrics.}}
Prior methods usually evaluate their IC performance on accuracy-based metrics.
Following ~\cite{tewel2022zerocap}, we use supervised metrics, $\ie$ metrics requiring human references, including BLEU-4 (B-4)~\cite{papineni2002bleu}, METEOR (M)~\cite{banerjee2005meteor}, CIDEr (C)~\cite{vedantam2015cider}, SPICE (S)~\cite{anderson2016spice} and RefCLIPScore (RefCLIP-S)~\cite{hessel2021clipscore}. 
RefCLIPScore measures the semantic similarity between references and predictions.
Besides, we also use an unsupervised metric, CLIPScore(CLIP-S)~\cite{hessel2021clipscore,tewel2022zerocap}. 
CLIPScore is a reference-free metric measuring the similarity between an image and the corresponding caption, which is the most critical metric for zero-shot IC~\cite{tewel2022zerocap}.
Another important performance of IC is to evaluate the diversity of generated captions. 
We follow~\cite{tewel2022zerocap, chen2021human} and use three metrics, Vocab~\cite{tewel2022zerocap}, Self-CIDEr(S-C)~\cite{wang2019describing} and Div-n~\cite{aneja2019sequential}. 
Vocab is the vocabulary size of all generated captions on the testing set, reflecting the word richness of different methods.
Self-CIDER and Div-n are popular diversity metrics based on pairwise similarities between captions.

{\bf{Quantitative Results}}
The quantitative results are reported in Table~\ref{Table:MSCOCO} based on the MSCOCO dataset. 
Obviously, for supervised metrics B-4, M, C, and S, zero-shot methods without any fine-tuning including ZeroCap and ours, lag behind other supervised methods.
This makes sense because supervised methods trained on MSCOCO can benefit domain bias, which means training and testing sets are labeled by the same annotators and thus have similar caption style.
However, as we discuss before, IC should not have the standard answers.
On the other hand, training or fine-tuning on one dataset does help models produce captions similar to the training data, but limits the word vocabulary size and thus decrease the diversity of generated captions.
Specifically, shown in the \textit{Diversity} column, even compared with existing supervised methods, Div-BS~\cite{wang2017diverse}, AG-CVAE~\cite{vijayakumar2018diverse}, POS~\cite{deshpande2019fast}, and ASG2Caption~\cite{chen2020say} that focus on improving the diversity, our method surpasses them with a large margin. 
Furthermore, our framework gets comparable and superior performance compared with SOTA methods on semantic-related metrics, RefCLIPScore and CLIPScore, respectively, indicating that our method can generate high image-matching caption. 

\begin{figure*}[!t]
	\centering 
	\includegraphics[width=160mm]{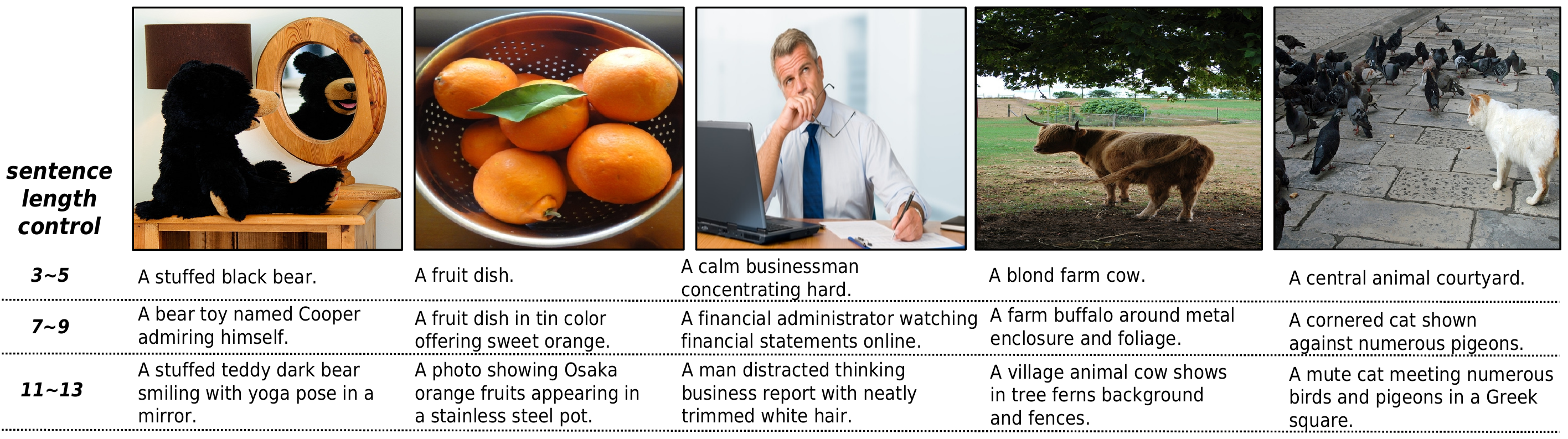}
	\vspace{-3mm}
	\caption{Examples of {\bf{length controlling}} by ConZIC. Given one image, we show the generated captions controlled by three different pre-defined lengths. Empirically, short captions are more global and generally describe the most salient object in images, while long captions talk about more visual details.}
	\label{Fig:length_examples}
	\vspace{-4mm}
\end{figure*}

Moreover, Table.~\ref{Table:sketchyCOCO} reports the zero-shot performance on sketch-style images (SketchyCOCO caption dataset) compared with previous SOTA methods. 
Our framework outperforms supervised methods (trained on MSCOCO) on most of the metrics because of the domain gap between MSCOCO and SketchyCOCO. 
Meanwhile, we surpass another zero-shot method ZeroCap on all metrics.

{\bf{Qualitative Results.}}
As shown in Figs.~\ref{fig1} and \ref{Fig1:model}, our framework can produce accurate and diverse captions with more words and abundant sentence patterns. 
More examples can be seen in Appendix ~\ref{appendixD}.

\subsection{Evaluation on controllable IC tasks}
\label{evaluation on CIC tasks}
We have considered 4 controllable tasks. The first two, $\ie$, length and infilling, are classifier-free. The last two, $\ie$, style and parts-of-speech, rely on an off-the-shelf classifier. 
We will detail each control task as follows.

\begin{figure}[!tb]
	\centering 
	\includegraphics[width=80mm]{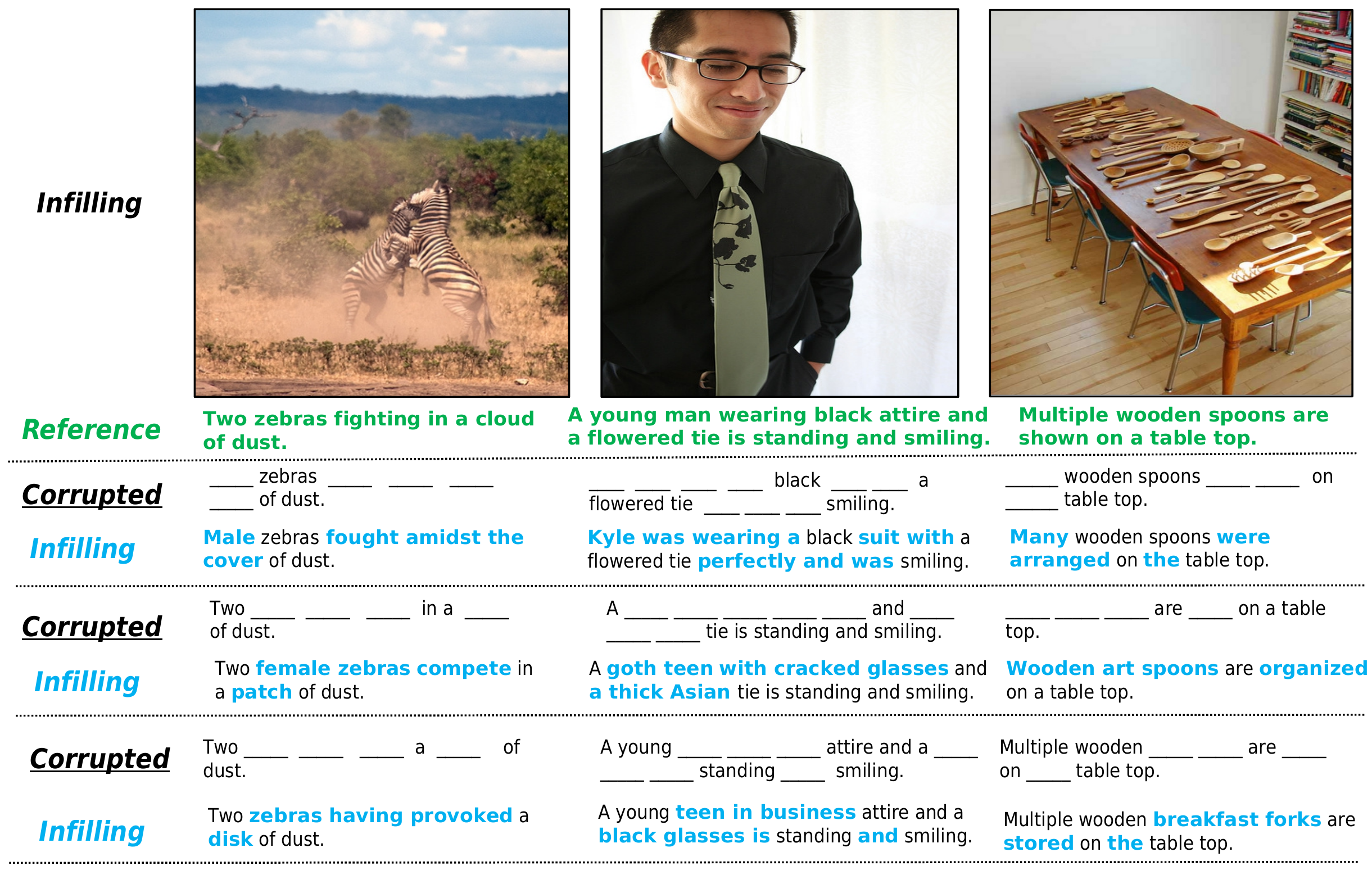}
	\vspace{-2mm}
	\caption{Examples of {\bf{infilling task}} by ConZIC. Given an image with a reference, we randomly corrupt some words. ConZIC infills these blanks to generate reasonable descriptions, where the infilling words are highlighted in blue. 
% 	Three different corrupted-infilling pairs results illustrate the robustness of ConZIC.
}
	\label{Fig:infilling_examples}
\vspace{-4mm}
\end{figure}

\begin{table}[tb!]
\begin{center}
  \resizebox{0.45\textwidth}{!}{
  \begin{tabular}{c|ccc|ccc}
    \toprule[2pt]
     & \multicolumn{3}{c|}{Noun} &\multicolumn{3}{c}{Verb}\\ 
    Method & B-1($\uparrow$) & WSim($\uparrow$)& BSim($\uparrow$) & B-1($\uparrow$) & WSim($\uparrow$)& BSim($\uparrow$) \\\hline
    ZeroCap~\cite{tewel2022zerocap} &0.00 &0.001 &0.11 &0.00 &0.001 &0.10 \\
    \textbf{ConZIC} &\textbf{0.37} &\textbf{0.39} &\textbf{0.52} &\textbf{0.25} &\textbf{0.46} &\textbf{0.50} \\
    \bottomrule[2pt]
  \end{tabular}}
  \end{center}
  \vspace{-6mm}
  \caption{\small{Results of one word infilling task on MSCOCO dataset.}}
\label{Table:infilling}
\vspace{-5mm}
\end{table}

{\bf{Length.}}
Length controlling means that we need to set a preferred length of captions.
For our framework ConZIC doing length-control IC, we just need to set the initial length of the caption without an extra classifier.
We do this experiment on MSCOCO.
Considering that the average sentence length of annotated captions in MSCOCO is about 10, we have tried 4 different lengths: 5, 8, 12, and 15.

The qualitative results are illustrated in Fig.~\ref{Fig:length_examples}.
Empirically, given a target length, ConZIC can generate accurate descriptions with a length within $\pm$2 due to the WordPiece-based tokenizer of BERT~\cite{wang2019bert}.
To understand the effect of  iteration steps and caption length in ConZIC, Figs.~\ref{length_clipscore} and \ref{length_refclipscore} show the change of CLIPScore and RefCLIPScore with the increase of iteration steps, with results of ZeroCap as a baseline.
These two scores increase with iterative updates, demonstrating the effectiveness of polishing mechanism.
Compared results from Fig.~\ref{Fig:length_examples} and Fig.~\ref{length_clipscore}, we observe that longer sentence generally contains more details of images, which facilitate a higher CLIPScore.
For RefCLIPScore in Fig.~\ref{length_refclipscore}, we find that
Length $12$ and Length $8$ have similar better results than Length $5$ and Length $15$.
We attribute it to the fact that the average length of caption in MSCOCO is about 10, and RefCLIPScore evaluates the similarity between generated and reference captions.

% \begin{table}[tb!]
% \begin{center}
%   \resizebox{0.45\textwidth}{!}{
%   \begin{tabular}{c|cccc}
%     \toprule[2pt]
%      & \multicolumn{4}{c}{Positive} \\
%     Methods & B-3($\uparrow$) &  M($\uparrow$) & CLIP-S($\uparrow$) &Acc($\uparrow$)\\ \hline
%     StyleNet ~\cite{gan2017stylenet} &12.1 &12.1 &- &45.2  \\
%     MSCap ~\cite{guo2019mscap} &16.2 &16.8 & - &92.5   \\
%     MemCap ~\cite{zhao2020memcap} &\textbf{17.0} &\textbf{16.6} &- &96.1 \\
%     \textbf{ConZIC} &1.89 &5.39 &\textbf{0.99} &\textbf{97.2} \\\hline
%     & \multicolumn{4}{c}{Negative} \\\hline
%     StyleNet ~\cite{gan2017stylenet}  &10.6  & 10.9 &- &56.6 \\
%     MSCap ~\cite{guo2019mscap}  &15.4 &16.2 &- &93.4  \\
%     MemCap ~\cite{zhao2020memcap}  &\textbf{18.1} &\textbf{15.7} &- &98.9 \\
%     \textbf{ConZIC}  &1.78 &5.54 &\textbf{0.97} &\textbf{99.1}  \\
%     \bottomrule[2pt]
%   \end{tabular}}
%   \end{center}
%   \vspace{-5mm}
%   \caption{Sentiments controlled image captioning ($\ie$ positive, negative) performance comparisons on the Senticap dataset. Acc is style classification accuracy.}
%   \label{Table:senticap}
%   \vspace{-4mm}
% \end{table}

{\bf{Infilling.}}
Given human-annotated caption with parts of words absent, infilling task targets at infilling suitable words conditioning on the image content.
As illustrated in Fig.~\ref{Fig:infilling_examples}, we consider this task as a special controllable IC task since we need to generate texts conditioning not only on the image content but also on the fixed left and right context. 
Most existing IC models, such as ZeroCap, can not do this task since the autoregressive LM they used can generate words only based on the left context.
On the contrary, ConZIC does this task without using a classifier since its LM, $\ie$ Gibbs-BERT, is modeled on bidirectional attention.

Firstly, we conduct quantitative experiments based on the setting where only one word is absent.
Specifically, we randomly mask one verb or noun in MSCOCO reference captions, and then require ConZIC to infill it.
Regarding the original word as the ground truth, we choose BLEU-1 (B-1) as the metric to evaluate the accuracy.
However, many other words are also suitable for this position (diversity).
Therefore, we use two metrics to measure the semantic similarity between predicted and reference words: WordNet word similarity (WSim)~\cite{pedersen2004wordnet}, and BERT embedding cosine similarity (BSim)~\cite{wang2019bert}. 
Results are shown in Table~\ref{Table:infilling}, where ConZIC outperforms ZeroCap by a large margin.
Moreover, as shown in Fig.~\ref{Fig:infilling_examples}, we provide some examples of infilling results by ConZIC, where for each image and reference caption, we prepare three randomly corrupted queries and ask ConZIC to infill words.
See Appendix ~\ref{appendixE} about the results of masking more words.

\begin{table}[tb!]
\begin{center}
  \resizebox{0.45\textwidth}{!}{
  \begin{tabular}{c|c|cccc}
    \toprule[2pt]
    \multicolumn{6}{c}{Positive} \\\hline
    \multicolumn{2}{c|}{Metrics} & B-3($\uparrow$) &  M($\uparrow$) & CLIP-S($\uparrow$) &Acc($\uparrow$)\\ \hline
    \multirow{3}{*}{Supervised}&StyleNet~\cite{gan2017stylenet} &12.1 &12.1 &- &45.2  \\
    &MSCap ~\cite{guo2019mscap} &16.2 &16.8 & - &92.5   \\
    &MemCap ~\cite{zhao2020memcap} &\textbf{17.0} &\textbf{16.6} &- &96.1 \\\hline
    Zero Shot&\textbf{ConZIC} &1.89 &5.39 &\textbf{0.99} &\textbf{97.2} \\\hline
    \multicolumn{6}{c}{Negative} \\\hline
    \multirow{3}{*}{Supervised}&StyleNet ~\cite{gan2017stylenet}  &10.6  & 10.9 &- &56.6 \\
    &MSCap ~\cite{guo2019mscap}  &15.4 &16.2 &- &93.4  \\
    &MemCap ~\cite{zhao2020memcap}  &\textbf{18.1} &\textbf{15.7} &- &98.9 \\\hline
    Zero Shot &\textbf{ConZIC}  &1.78 &5.54 &\textbf{0.97} &\textbf{99.1}  \\
    \bottomrule[2pt]
  \end{tabular}}
  \end{center}
  \vspace{-7mm}
  \caption{Sentiment controlled image captioning ($\ie$ positive, negative) performance comparisons on the SentiCap dataset. Acc is style classification accuracy.}
  \label{Table:senticap}
  \vspace{-5mm}
\end{table}

\begin{figure}[!t]
	\centering 
	\includegraphics[width=80mm]{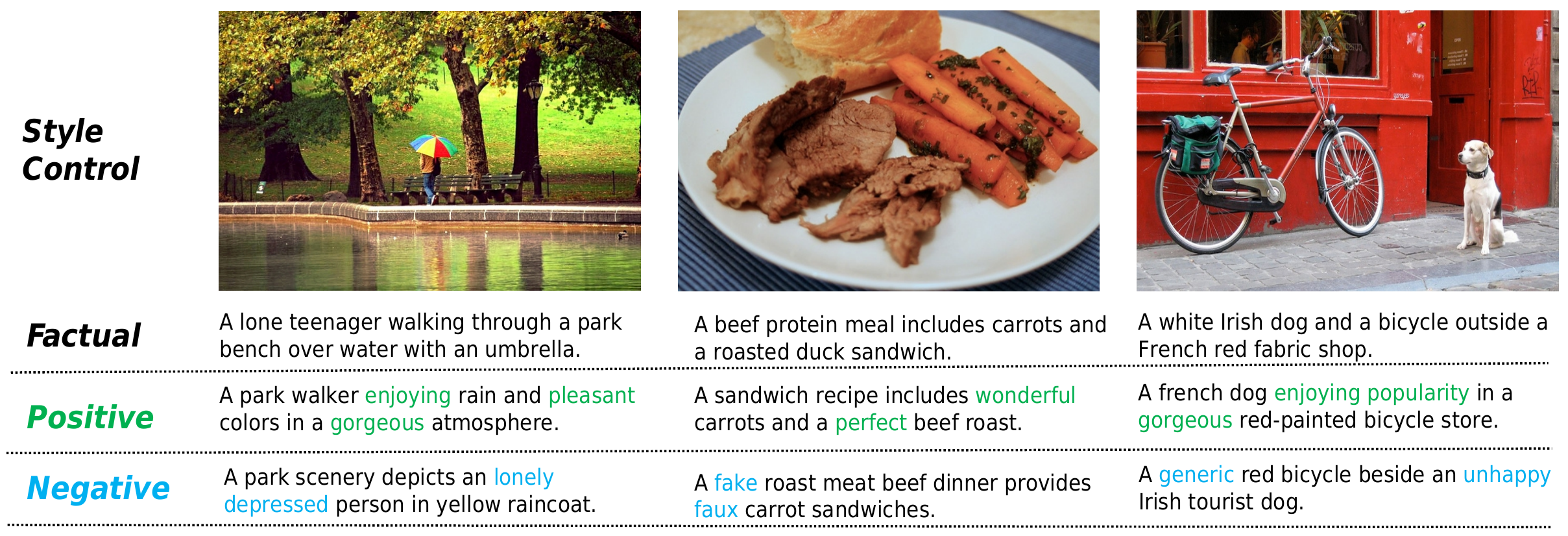}
\vspace{-3mm}
	\caption{Examples of {\bf{sentiment controlled task}} by ConZIC. Factual is the captions without controlling generated by our proposed framework (ConZIC without discriminator of control signal). Positive and Negative are the stylized image captioning results of ConZIC, where sentiment-related words are highlighted in green and blue, respectively.}
	\label{Fig:style_examples}
\vspace{-4mm}
\end{figure}

{\bf{Style.}}
Given an image with a specific linguistic style, $\eg$, positive, negative, romantic, or humorous, style-controlled task aims to generate corresponding descriptions.
For this task, ConZIC needs a pre-trained classifier to discriminate the style of captions.
Currently, we use SentiwordNet ~\cite{baccianella2010sentiwordnet} for sentiment (positive or negative) controlling and use TextCNN ~\cite{guo2019improving} for romantic-humorous controlling.

Firstly, we evaluate the sentiment-controlled capability of ConZIC on the SentiCap dataset. The quantitative results are shown in Table~\ref{Table:senticap}.
As baselines, StyleNet~\cite{gan2017stylenet}, MSCap~\cite{guo2019mscap}, and Memcap~\cite{zhao2020memcap} achieve the SOTA performance on this task in \textit{supervised} way, resulting in a higher B-3 and M.
Following them, we test the accuracy (Acc) of ConZIC, which is obtained by feeding the generated captions into a sentiment classifier, where ConZIC are higher.
Furthermore, we use CLIP-S to evaluate the correlation between image and caption whose results demonstrate that the captions generated by ConZIC are highly correlated to the images.
Since three baselines do not provide public codes, we cannot evaluate them on CLIP-S but just report other three metrics from the paper.
For better visualization of ConZIC for sentiment controlling, Figs.~\ref{fig1b}, \ref{Fig1:model}, and \ref{Fig:style_examples} shows multiple results.
Results and analysis on the FlickStyle10k about romantic-humorous styles are in Appendix ~\ref{appendixE}.

{\bf{Parts-of-speech (POS).}}
The task of POS-control IC is to generate captions which match given POS tags.
For example, the POS of caption \textit{a cat sitting in the bed} is \textit{DET NOUN VERB ADP DET NOUN}.
For this task, we use the POS classifier developed  in~\cite{bird2006nltk}.
Considering a fact that human generally describes an image under some common templates, like ``somebody/something doing something at someplace'', we design a POS tag sequence as \textit{DET ADJ/NOUN NOUN VERB VERB ADV ADP DET ADJ/NOUN NOUN NOUN}.
Under such template, we report the results of ConZIC in Table~\ref{Table:POS}.
Clearly, Our method can achieve a high accuracy under this POS tag template, but METEOR (M), CIDEr (C), CLIPScore (CLIP-S) gets slightly lower because not all images are suitable for this POS tags. 
We also visualize some examples in Fig.~\ref{Fig:POS_examples}.
More analysis and results are discussed in Appendix ~\ref{appendixE}.

\begin{figure}[!t]
	\centering 
	\includegraphics[width=83mm]{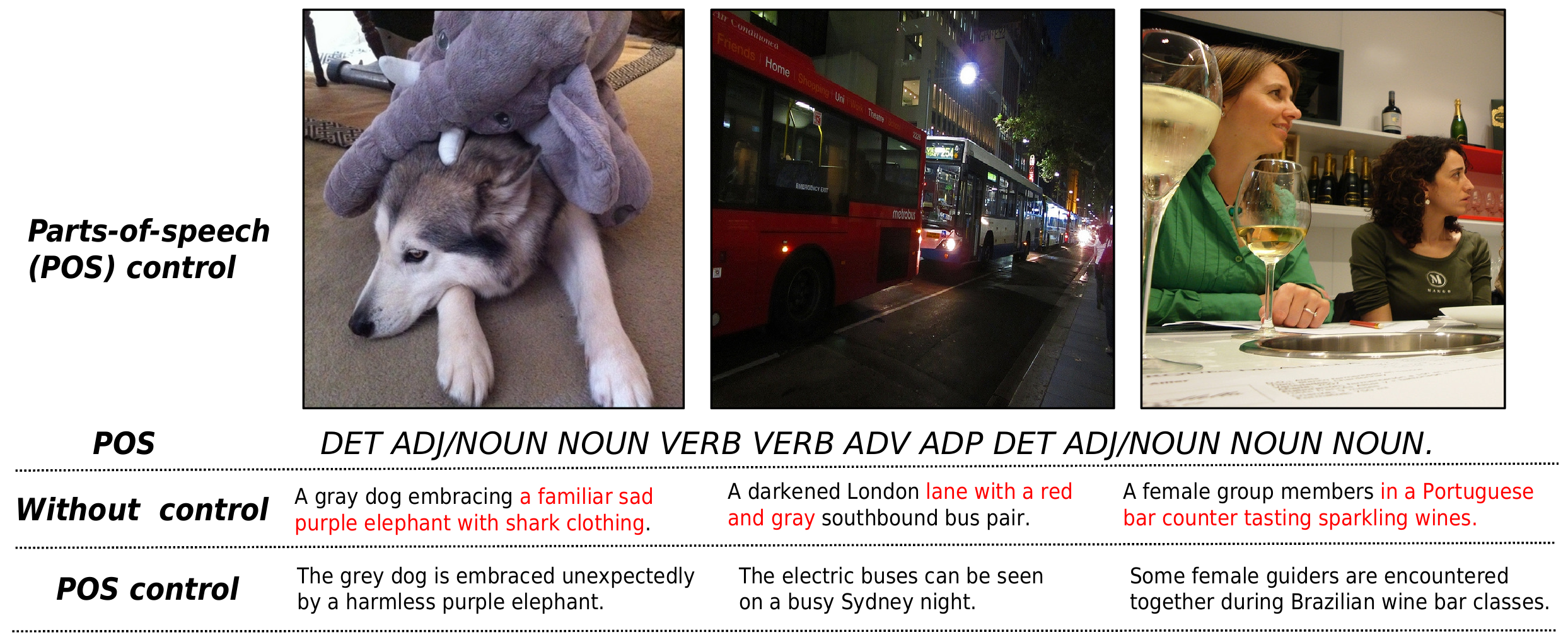}
	\vspace{-6mm}
	\caption{Examples of {\bf{parts-of-speech controlling}} (POS) by ConZIC. \textit{DET ... NOUN} is the predefined POS template. ``Without POS" denotes image captioning results without controlling by our proposed framework (ConZIC without discriminator of control signal) where words in color represents they do not satisfy the POS. ``POS control" denotes the controllable results of ConZIC.}\label{Fig:POS_examples}
\vspace{-4mm}
\end{figure}

\begin{table}[tb!]
\begin{center}
  \resizebox{0.45\textwidth}{!}{
    \begin{tabular}{c|cccc}
    \toprule[2pt]
    Parts-of-speech & M($\uparrow$) &C($\uparrow$) & CLIP-S($\uparrow$)&Acc($\uparrow$)\\ \hline
    without POS &11.54 &12.84 &1.01 &15.54 \\
    with POS &8.25 &10.89 &0.96 &83.36 \\
    \bottomrule[2pt]
    \end{tabular}}
  \end{center}
  \vspace{-6mm}
  \caption{\small{Results of parts-of-speech control on MSCOCO. Pre-defined POS tags is \textit{DET ADJ/NOUN NOUN VERB VERB ADV ADP DET ADJ/NOUN NOUN NOUN.}}}
  \label{Table:POS}
  \vspace{-4mm}
\end{table}
\subsection{Evaluation on generation speed}
\label{Evaluation on Time-consuming}
The generation speed is significant for zero-shot IC.
For ZeroCap and ConZIC, empirically, we find that the speed is mostly related to sentence length.
Fig.~\ref{length_speed} shows the generation speed of ConZIC and ZeroCap with different sentence lengths, evaluated on MSCOCO. 
Our method is around 5 times faster than ZeroCap with 15 iterations, which attributes to our proposed sampling-based Gibbs-BERT. 

\vspace{-2mm}
\section{Conclusion and future work}
\vspace{-1mm}
In this paper, we propose a flexible and efficient framework for controllable zero-shot IC, named ConZIC.
Firstly, by discovering the relation between MLMs and Gibbs sampling, we develop a new sampling-based language model called Gibbs-BERT. 
Compared with widely used autoregressive models, Gibbs-BERT has flexible generation order, bringing the self-correct capability by bidirectional attention.
% and faster and more diverse generation
To integrate Gibbs-BERT with the CLIP for image-text matching and the pre-trained discriminator for controlling, ConZIC can realize better zero-shot IC with/without control signals.
However, as our analysis of failure examples on ConZIC and ZeroCap in Appendix ~\ref{appendixF}, the study of zero-shot IC is still in its infancy, leaving a large space to go further.
For example, ConZIC and ZeroCap often ignore small targets in the image, which may be alleviated by a detector for capturing small targets in the image.
Besides, developing more appropriate metrics, especially for controllable IC, rather than using those supervised metrics, is also important for the development of zero-shot IC.

%%%%%%%%% REFERENCES
{\small
%\bibliographystyle{ieee_fullname}
% \bibliography{egbib}

}

\clearpage
\appendix
% \section{*Appendix}
\begin{table*}[!htb]
\centering
\resizebox{1\textwidth}{!}{
  \begin{tabular}{c|c|c}
    \toprule[1.5pt]
     & Novel object captioning & ZeroCap or ConZIC \\ \hline
     generalization ability & seen objects IC $\rightarrow$ unseen objects IC & large pretrained models $\rightarrow$ image captioning task \\
     &  (with limited background/image styles) & (with no limitations on objects/background/image styles)\\ \hline
     well designed image-caption pairs & needed for training or fine-tuning & no need \\ \hline
     extra knowledge & object taggers & no need \\
    \bottomrule[1.5pt]    
  \end{tabular}}  
  \caption{Comparisons on problem scenarios of NOIC and our ZIC.}
  \label{compare to related work}
\end{table*}

\section{Discussion about NOIC and ZIC}
\label{appendixA}
In this section, we will discuss the difference between novel object image captioning (NOIC) and zero-shot image captioning (ZIC), brief comparisons are shown in Table \ref{compare to related work} and details are as follows:

%Detailed explanations to Table \ref{compare to related work} are as follows:
%For clear demonstration, detailed comparisons are discussed below.

$\bullet$ {\bf\textit{Generalization among objects vs. among tasks.}} NOC aims to generalize image captioning (IC) models to ``novel objects'' not presented in the training images.
This means both training and testing tasks are IC.
{\bf{By contrast}}, the ``zero-shot'' concept in our work (and most related work in our paper) comes from GPT-3, referring to {\textit{applying large pre-trained models}} (trained with no specific task) {\textit{for downstream IC tasks with no task-specific fine-tuning}}. 

$\bullet$ {\bf\textit{With vs. without curated training image-caption pairs.}}
NOC models are often trained on well-designed image-caption pairs of seen objects. 
Hence, different dataset splits are often considered to perform evaluation.
{\bf{By contrast}}, ConZIC is free of well-designed image-caption pairs to perform training or even fine-tuning.

$\bullet$ {\bf\textit{With vs. without extra knowledge.}}
NOC methods often learn the relations between objects and extra taggers, such as attributes and class embeddings.
Then, these relations are generalized to unseen objects
by various techniques. {\bf{By contrast}}, ConZIC utilizes the knowledge from large pre-trained models and thus is free of extra information.

\section{Algorithm of Gibbs-BERT}
\label{appendixB}
After randomly choosing the generation order, Gibbs-BERT starts from a full noisy sentence ($\eg$, all [{\textit{MASK}}] tokens).
At each iteration, Gibbs-BERT progressively samples each word by putting [{\textit{MASK}}] at this position and then selecting the top-$1$ word from the predicted word distribution over the vocabulary by BERT.
The result of $t$-th iteration is the initialization of the $(t+1)$-th iteration. The pseudo-code is shown in algorithm.~\ref{alg::Gibbs-BERT}

\begin{algorithm}[!htb]
\caption{Algorithm of Gibbs-BERT.} % 名称
\label{alg::Gibbs-BERT}
\KwData{initial sentence:\small{$\xv_{<1,n>}^0=(x_1^0,...,x_n^0)$}\;
iterations=\small{$T$}, candidates=\small{$K$}\; position sequence \small{$P=\textit{Shuffle}([1,...,n])$\;}}
\KwResult{the final sentence: \small{$\xv_{<1,n>}^T=(x_1^T,...,x_n^T)$;}}
\For{iteration \small{$t \in [1,...,T]$}}
{ 
    state: \small{$\xv_{<1,n>}^{t-1}=(x_1^{t-1},...,x_n^{t-1})$}\; 
    \For{position $i \in P$}
    {
        1. Replace \small{$x_i^{t-1}$} with \small{[\textit{MASK}]}\;
        2. Predict the word distribution over vocabulary by BERT: 
        \small{$p(x_i|\xv_{-i}^{t-1})$}\;
        3. Sample \small{$x_i$} from distribution \small{$p(x_i|\xv_{-i}^{t-1})$}\;
        4. Replace \small{$x_i^{t-1}$} with \small{$x_i^{t}$}\;
    }
    state: \small{$\xv_{<1,n>}^{t}=(x_1^{t},...,x_n^{t})$}\; 
}
\end{algorithm}

\begin{figure*}[!tb]
	\centering 
	\includegraphics[width=151mm]{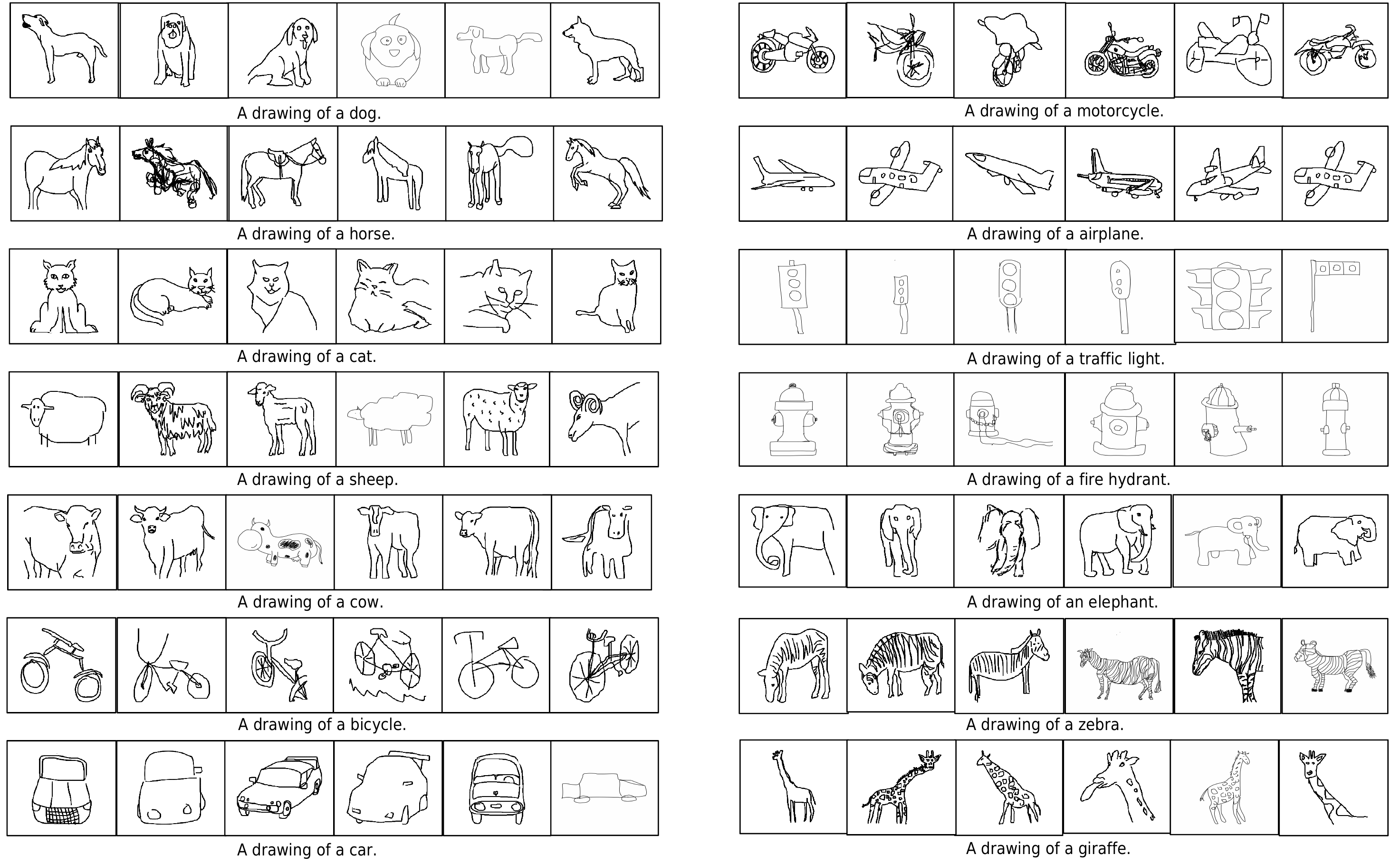}
        \vspace{-3mm}
	\caption{Examples of SketchyCOCO caption benchmark
	}\label{fig:sketchycoco}

\end{figure*}

\section{SketchyCOCO caption benchmark}
\label{appendixC}
SketchyCOCO caption is a small sketch-style image captioning benchmark based on SketchyCOCO, including 14 classes, as shown in Fig.~\ref{fig:sketchycoco}.
SketchyCOCO is not an image captioning dataset since it only has the classification label.
we construct the captioning benchmark through a text prompt ``A drawing of a [CLASS]'', where [CLASS] is the class name.  

\section{More generation examples}
\label{appendixD}
\begin{figure*}[!tb]
	\centering 
	\includegraphics[width=151mm]{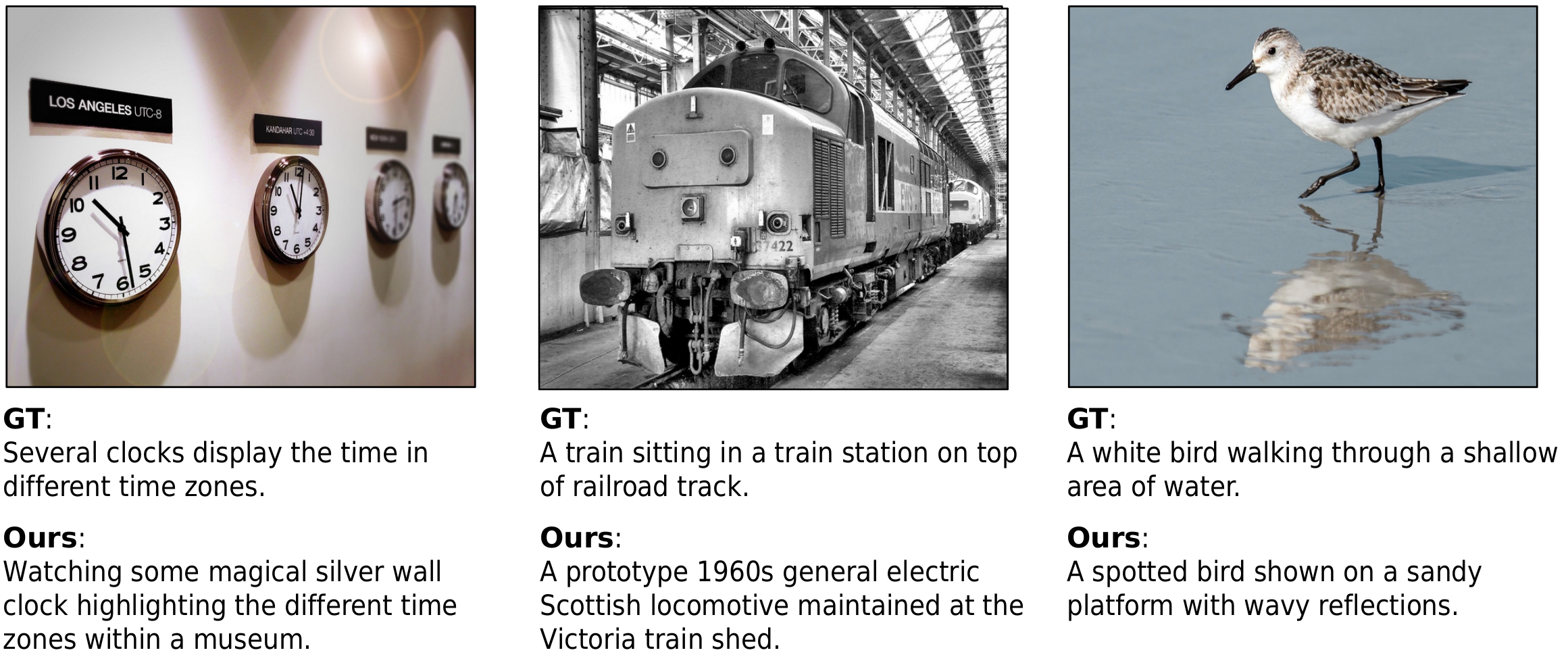}
	\caption{Comparision with groundtruth on MSCOCO caption.
	}\label{fig:gt}
\end{figure*}

\textbf{Comparison with Ground-Truth.}
As shown in Fig.~\ref{fig:gt}, due to the zero-shot nature, caption generations of our method are different from MSCOCO ground-truth.
our method has shown significant differences with ground-truth in syntactic(sentence patterns) and semantic(diverse words).  

\textbf{Diverse generation compared with ZeroCap.}
Comparison results on diverse caption generation are shown in Fig.~\ref{fig:diversity}.
ZeroCap generates diverse captions by beam search, which can result in a similar sentence pattern with respect to mode collapse.
In contrast, our method can produce multiple captions related to the same image by shuffling the word generation order, which has shown strong performance in syntactic and semantic diversity.

\begin{table}[tb!]
\begin{center}
  \resizebox{0.45\textwidth}{!}{
  \begin{tabular}{c|ccc}
    \toprule[1.5pt]
    Diversity Metrics & S-C($\uparrow$) & Div-1($\uparrow$) & Div-2($\uparrow$)\\ \hline
    Zerocap &0.63 &0.40 &0.56   \\
    \textbf{Ours} &\textbf{0.95} &\textbf{0.63} & \textbf{0.84}  \\
    \bottomrule[1.5pt]
  \end{tabular}}
  \end{center}
\vspace{-6mm}
\caption{\small{Length controlled diversity metrics of our method on MSCOCO. we select the best-1 caption on each length and then compute diversity metrics conditioning on these four captions. }}
 \label{Table:length}
 \vspace{-4mm}
\end{table}

\begin{table}[tb!]
\begin{center}
  \resizebox{0.45\textwidth}{!}{
    \begin{tabular}{c|cccc}
    \toprule[2pt]
    Corrupted Ratio &B-4($\uparrow$) & M($\uparrow$) & CLIP-S($\uparrow$)\\ \hline
    0.25 &60.69 &44.99 &0.83 \\
    0.50 &26.08 &29.12 &0.89 \\
    0.75 &8.06 &17.60 &0.93 \\
    \bottomrule[2pt]
    \end{tabular}}
  \end{center}
  \vspace{-6mm}
  \caption{\small{Results of multiple-words-infilling task.}}
  \label{Table:multiple-words-infilling}
  \vspace{-4mm}
\end{table}

\textbf{Results on various image styles and world knowledge.}
As shown in Fig.~\ref{fig:worldknowledge1} and Fig.~\ref{fig:worldknowledge2}. Our method performs well in various image styles, $\eg$ natural images, medical images, oil paintings and cartoon images. Besides, our method is proven to have efficient application in images with abundant world knowledge, $\eg$ medical, geography, celebrity, and artworks.

\section{Controllable tasks}
\label{appendixE}
\subsection{Diversity of length control}
Table~\ref{Table:length} has reported diversity performance where we select the best-1 caption on each length and then compute diversity metrics on these four lengths. Our method surpasses ZeroCap by a large margin.
\subsection{Infilling tasks}

We have conducted experiments on one-word-infilling and multiple-word-infilling.

\textbf{One-word-infilling.}
% Firstly, we choose one human-annotated caption $X=\{x_i\}_1^{n}$ in MSCOCO dataset. 
% Then, by utilizing an off-the-shelf parts-of-speech(POS) tagger, we randomly select a word $x_j$ whose POS is a noun/verb and view other words $x_{<j},x_{>j}$ as the context query. 
% For one-word-infilling, models need to infilling the most suitable word in this position given the context query. 
We randomly corrupt one verb/noun in the reference caption, and ask models to infill the most suitable word given other words. We use three metrics to evaluate the accuracy performance:
\textit{1)} BLEU-1(B-1) to measure unigram precision;
\textit{2)} Wordnet path similarity(WSim) which measures node distance in Wordnet.
Especially, this metric can only be computed between two words of the same POS. Therefore, we set WSim as 0 when the answer has a different POS from the reference word;
\textit{3)} BERT word similarity(BSim). We use cosine distance in BERT word embedding space, where words have similar semantics generally possess a low distance. Due to its autoregressive nature, ZeroCap can only take the left context into account, which limits its performance. The results are shown in Table.~\ref{Table:infilling}. 

\textbf{Multiple-word-infilling}
In contrast to one-word-infilling, we try to corrupt more words in reference caption, as illustrated in Fig.~\ref{Fig:infilling_examples}. Results with different corrupted ratios are shown in Table. \ref{Table:multiple-words-infilling}. We can see that results of a higher corrupted ratio are generally higher in CLIP-S and lower in other metrics.

\begin{figure*}[!tb]
	\centering 
	\includegraphics[width=151mm]{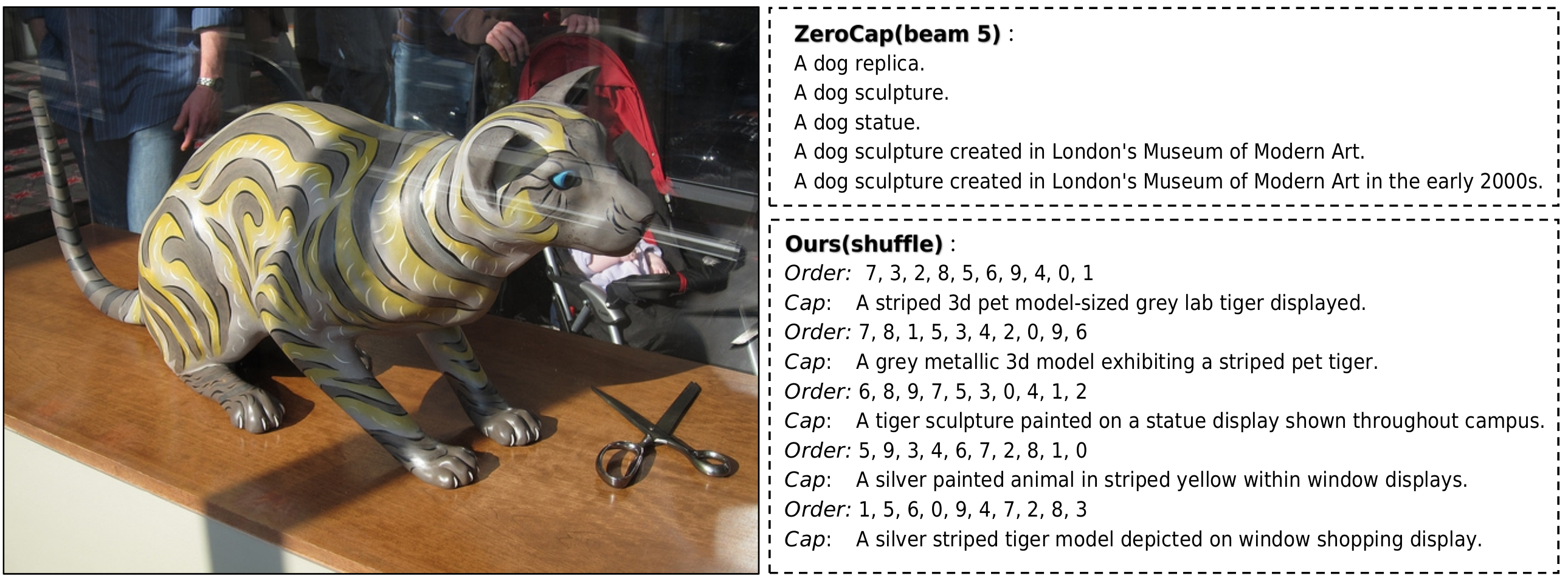}
	\caption{Diversity results compared with ZeroCap.
	}\label{fig:diversity}
\end{figure*}

\begin{figure*}[!tb]
	\centering 
	\includegraphics[width=151mm]{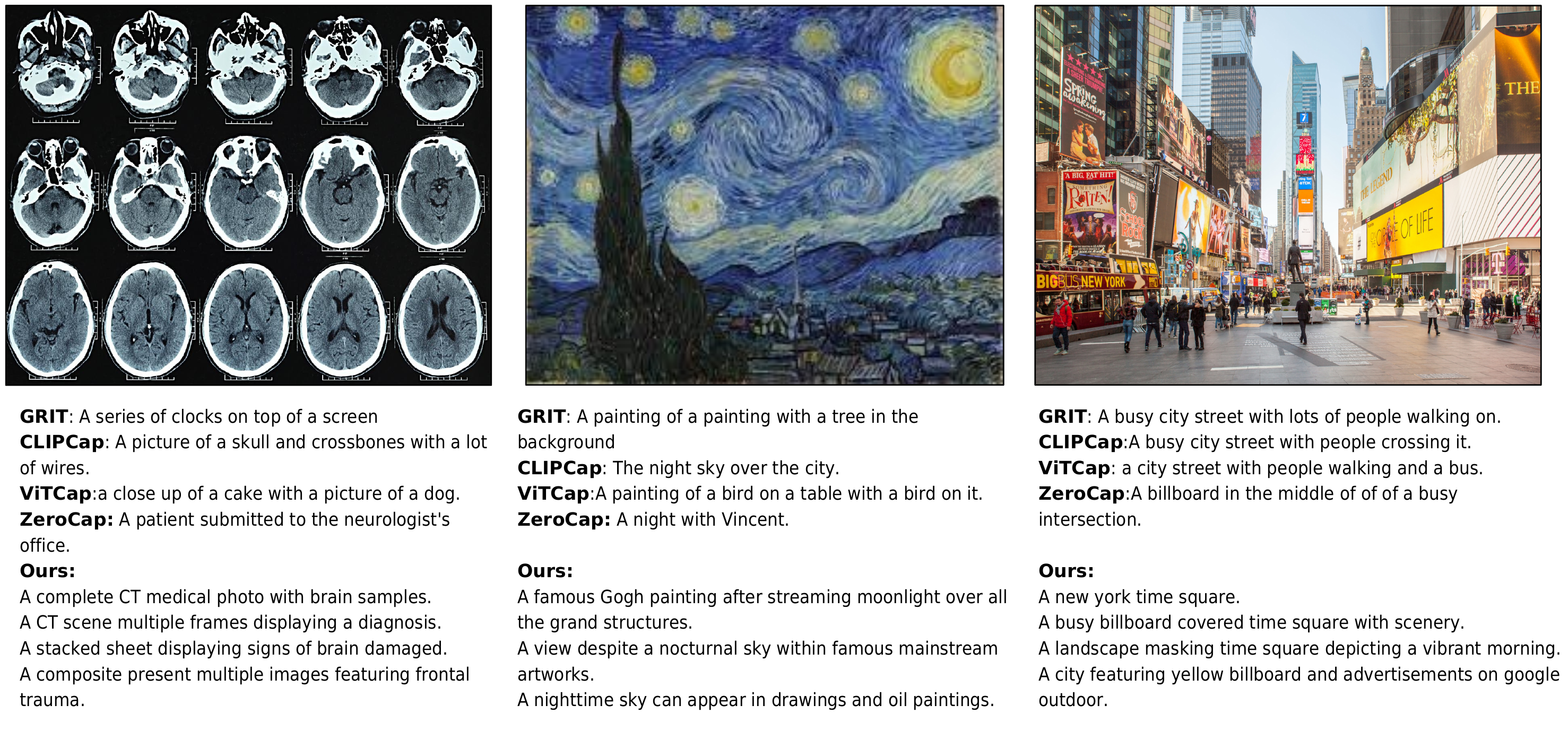}
	\caption{Results of various image styles.
	}\label{fig:worldknowledge1}
\vspace{-5mm}
\end{figure*}

\begin{figure*}[!tb]
	\centering 
	\includegraphics[width=151mm]{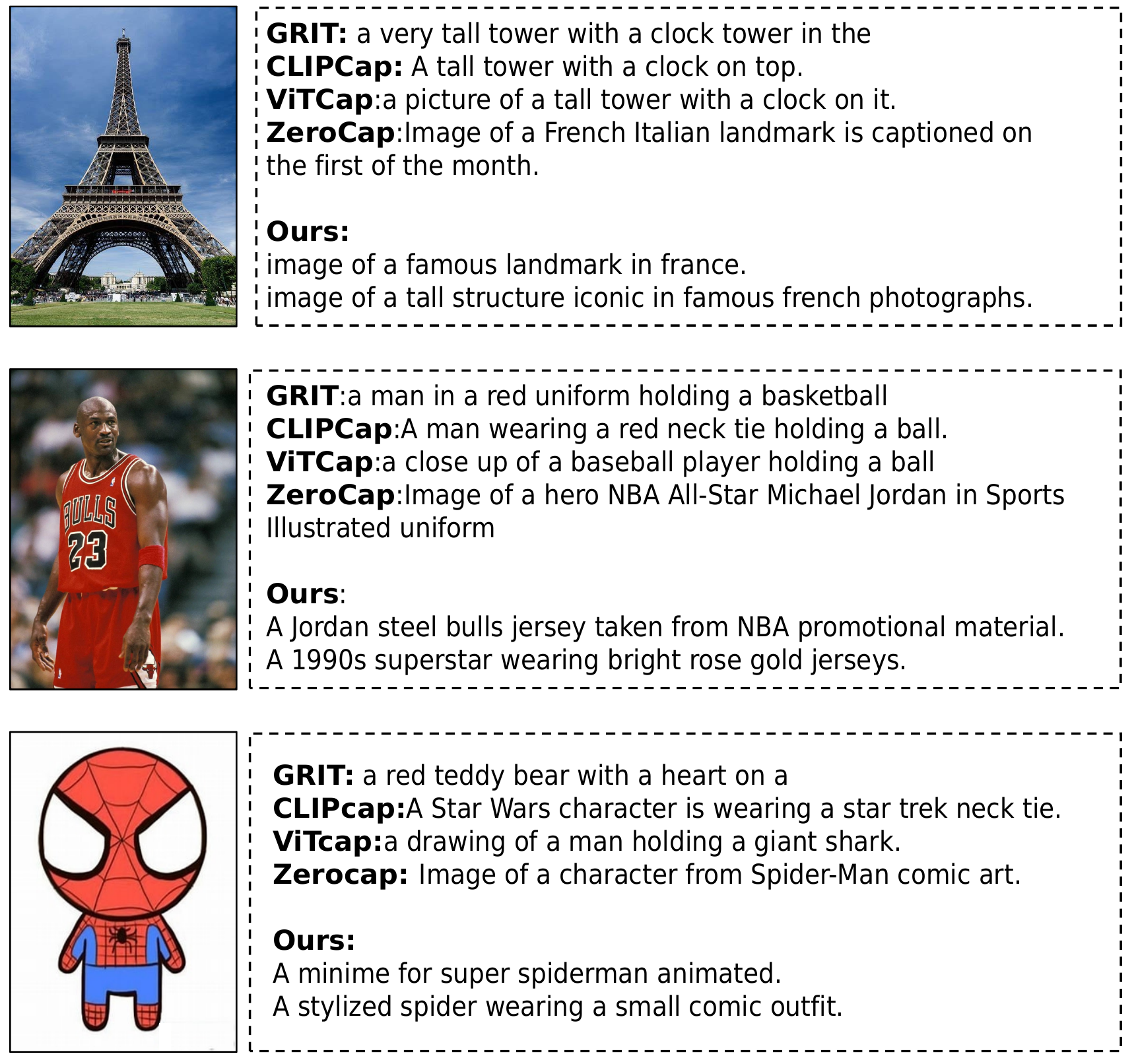}
	\caption{Results of images containing world knowledge.
	}\label{fig:worldknowledge2}
\vspace{-2mm}
\end{figure*}

\subsection{Humorous-Romantic control on FlickStyle10k}
Quantitative results are shown in Table. ~\ref{Table:flickstyle}.
As we can see, our method has comparable performance in producing captions in specific styles, $\ie$ romantic and humorous, as shown in \textit{Acc} column.
\begin{table*}[tb!]
\begin{center}
  \resizebox{0.9\textwidth}{!}{
  \begin{tabular}{c|cccc|cccc}
    \toprule[2pt]
    & \multicolumn{4}{c|}{Romantic} & \multicolumn{4}{c}{Humorous} \\
    Methods & B-3($\uparrow$) &  M($\uparrow$) & CLIP-S($\uparrow$) &Acc($\uparrow$) &B-3($\uparrow$) &M($\uparrow$) & CLIP-S($\uparrow$)  &Acc($\uparrow$)\\ \hline
    StyleNet  &1.5 &4.5 &- &37.8 &0.90  & 4.30 &- &41.9 \\
    MSCap  &2.0 &5.4 &- &88.7 &1.90 &5.30 &- &91.3  \\
    MemCap &\textbf{4.0} &\textbf{7.7} &- &91.7 &\textbf{4.0}  &\textbf{7.20} &- &\textbf{97.1} \\\hline
    \textbf{Ours} &1.2 &6.1 &\textbf{1.02} &\textbf{96.3} &1.2 &6.1 &\textbf{1.02} &91.4  \\
    \bottomrule[2pt]
  \end{tabular}}
  \end{center}
  \vspace{-5mm}
  \caption{Stylized image captioning($\ie$ romantic, humorous) performance comparisons on the Flickstyle8k dataset.}
\label{Table:flickstyle}
\end{table*}

\subsection{Parts-of-speech controlling}
we have tried another POS sequence, \textit{ADP DET ADJ/NOUN NOUN NOUN DET ADJ/NOUN NOUN VERB VERB ADV }. Results are shown in Table.~\ref{Table:another pos}.

\begin{table}
\begin{center}
  \resizebox{0.45\textwidth}{!}{
    \begin{tabular}{c|cccc}
    \toprule[2pt]
    Parts-of-speech & M($\uparrow$) &C($\uparrow$) & CLIP-S($\uparrow$)&Acc($\uparrow$)\\ \hline
    without POS &11.54 &12.84 &1.01 &15.54 \\
    with POS &7.99 &9.29 &0.95 &86.20 \\
    \bottomrule[2pt]
    \end{tabular}}
  \end{center}
  \vspace{-6mm}
  \caption{\small{Results of parts-of-speech control on MSCOCO. Pre-defined POS tags is \textit{ADP DET ADJ/NOUN NOUN NOUN DET ADJ/NOUN NOUN VERB VERB ADV }}}
  \label{Table:another pos}
\end{table}

\section{Bad case analysis}
\label{appendixF}
As shown in Fig.~\ref{fig:diversity}, ZeroCap and our method both ignore the ``scissor'' around the ``tiger statue'', which means that how to control which image content to be described, in particular, small objects, is under-explored for zero-shot image captioning.

\end{document}